\documentclass[acmtog,nonacm]{acmart}

\makeatletter
\if@ACM@newfonts\else
  \usepackage{lmodern}
\fi
\makeatother

\usepackage{enumitem}
\usepackage{cleveref}
\setcopyright{none}
\newcommand{\para}[1]{\noindent\textbf{\textit{#1}}}

\begin{document}

\title{DragonMoCap: Monocular Video-based Motion Capture for Diverse Dragon Topologies}

\author{Zhongfei Qing}
\email{kinnyqing@tencent.com}
\orcid{0009-0008-3101-8349}
\affiliation{\institution{Tencent}
  \city{Shenzhen}
  \country{China}
}

\author{Xiaoxiang Zhao}
\email{charlixzhao@tencent.com}
\orcid{0009-0006-8826-591X}
\affiliation{\institution{Tencent}
  \city{Shenzhen}
  \country{China}
}

\author{Ling-Hao Chen}
\email{clh22@mails.tsinghua.edu.cn}
\orcid{0000-0002-2528-6178}
\affiliation{\institution{Tsinghua University}
  \city{Shenzhen}
  \country{China}
}

\author{Zhongang Cai}
\email{caiz0023@e.ntu.edu.sg}
\orcid{0000-0002-1810-3855}
\affiliation{\institution{Nanyang Technological University}
  \city{Singapore}
  \country{Singapore}
}

\author{Wenjia Wang}
\email{wwj2022@connect.hku.hk}
\orcid{0000-0003-0121-3852}
\affiliation{\institution{The University of Hong Kong}
  \city{Shenzhen}
  \country{China}
}

\author{Yifan Wu}
\email{wuyifan1@bu.edu}
\orcid{0009-0009-7612-3297}
\affiliation{\institution{Tencent}
  \city{Shenzhen}
  \country{China}
}

\author{Qianying Liu}
\email{2665227L@student.gla.ac.uk}
\orcid{0000-0003-0765-7243}
\affiliation{\institution{University of Glasgow}
  \city{Shanghai}
  \country{China}
}

\author{Wei Zeng}
\email{s-zw24@bza.edu.cn}
\orcid{0009-0000-0448-1057}
\affiliation{\institution{Tencent}
  \institution{Southern University of Science and Technology}
  \city{Shenzhen}
  \country{China}
}

\author{Bo Hu}
\authornotemark[2]
\email{corehu@tencent.com}
\orcid{0000-0002-6599-7249}
\affiliation{\institution{Tencent}
  \city{Shenzhen}
  \country{China}
}

\author{Songnan Li}
\authornotemark[1]
\email{sunnysnli@tencent.com}
\orcid{0000-0001-6931-4129}
\affiliation{\institution{Tencent}
  \city{Shenzhen}
  \country{China}
}

\makeatletter
\g@addto@macro\@authornotes{\stepcounter{footnote}\footnotetext{Corresponding author: Songnan Li (sunnysnli@tencent.com).}}
\g@addto@macro\@authornotes{\stepcounter{footnote}\footnotetext{Project lead: Bo Hu.}}
\makeatother

\renewcommand{\shortauthors}{Qing et al.}

\hypersetup{
  pdftitle={DragonMoCap: Monocular Video-based Motion Capture for Diverse Dragon Topologies},
  pdfauthor={Zhongfei Qing, Xiaoxiang Zhao, Ling-Hao Chen, Zhongang Cai, Wenjia Wang, Yifan Wu, Qianying Liu, Wei Zeng, Bo Hu, Songnan Li}
}

\begin{abstract}
This work explores monocular motion capture for dragon-type characters, which currently relies on manual keyframing due to the lack of high-quality datasets and parametric models. The primary obstacle lies in the extreme topological diversity across dragon rigs, where varying joint counts and naming conventions render traditional fixed-template regressors inapplicable. To address this, we introduce \emph{DragonMoCap}, a novel framework that treats the target rig as a dynamic input rather than a fixed prior. By deliberately scoping the framework to the most prevalent production topologies, we ensure high-fidelity results across diverse morphologies. The core of our approach is the design of \emph{language-grounded joint query tokens}, which establish a dual-property representation: ensuring intra-rig discriminability while maintaining cross-rig similarity. We further propose a non-parametric \emph{query-token retrieval} mechanism to enable zero-shot adaptation to unseen rigs without retraining. Extensive evaluations demonstrate that \emph{DragonMoCap} consistently outperforms HMR-style baselines across diverse dragon topologies. The practical utility of our framework is further evidenced by its integration into professional animation workflows, highlighting its potential for real-time industrial applications.
\end{abstract}




\begin{teaserfigure}
  \includegraphics[width=\textwidth]{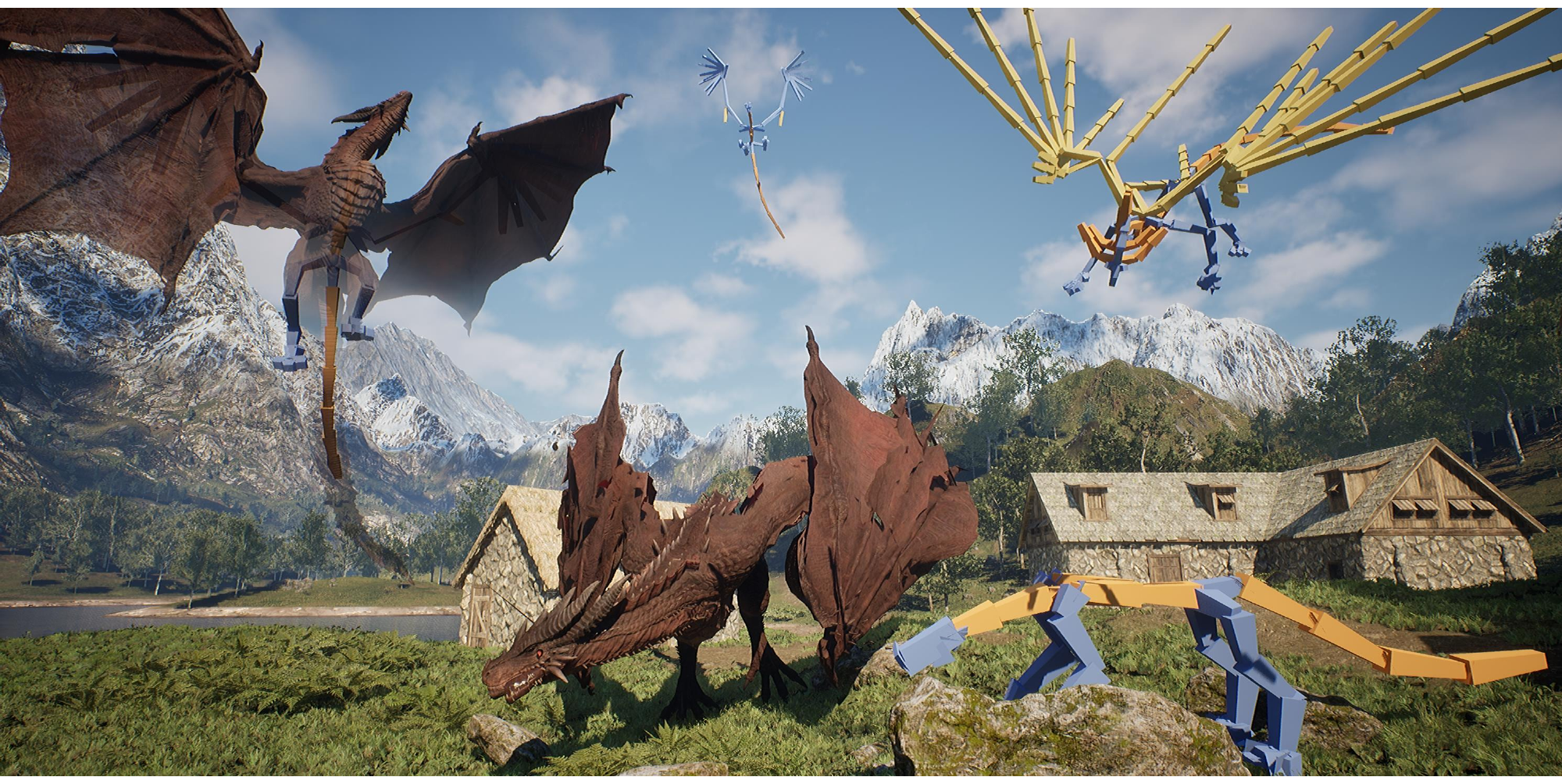}
  \vspace{-2em}
  \caption{\emph{DragonMocap} captures a dragon motion from a monocular video and retargets it to diverse dragon rigs (Bipedal or quadrupedal, with or without wings). }
  \label{fig:teaser}
\end{teaserfigure}

\maketitle

\section{Introduction}

Dragons occupy a pivotal position in modern digital content, ranging from AAA games to feature films. Despite their ubiquity, dragon animation remains a labor-intensive manual craft. Established pipelines that have commoditized human animation, such as large-scale mocap datasets and parametric models like SMPL(-X), fail to generalize to these complex fantasy creatures. This gap stems from two primary challenges. First, there is a critical scarcity of high-quality dragon motion data. Second, the lack of a canonical skeleton or consistent bone naming creates a significant topological barrier. Production rigs often feature highly diverse configurations, with joint counts ranging from 20 to over 60 and varying wing or leg structures. Consequently, a universal video-based motion capture system for dragons remains an elusive goal.

Existing monocular motion capture methodologies are intrinsically tied to fixed structural priors, such as the SMPL model, and typically regress a fixed-size output vector. This structural rigidity prevents these frameworks from being redirected to characters with different joint counts or hierarchical topologies. Specifically, the field lacks a video-based regressor capable of recovering high-fidelity motion directly from monocular frames when no clean 3D source is available. Furthermore, such a system must generalize to arbitrary dragon rigs without the need for expensive per-case finetuning or labor-intensive manual correspondences. As a result, introducing a topology-agnostic and flexible framework for dragon motion capture and synthesis is fundamental and meaningful.

To address these challenges, we introduce \emph{DragonMoCap}, the first video-based motion capture framework designed for dragon-type characters. Given a monocular input video and a user-provided rig, defined by its bone names and hierarchy, our method predicts per-joint rotations independently for each frame and concatenates the predictions into a motion sequence. To ensure high-fidelity results and rigorous evaluation, we deliberately scope the framework to the three most prevalent production topologies: winged quadrupeds, wingless quadrupeds, and winged bipeds (wyverns). This targeted approach allows the model to be trained and validated on a high-quality dataset of professional animations while maintaining broad applicability across the most common fantasy morphologies.

The core design principle of \emph{DragonMoCap} is to treat the target rig as a dynamic input rather than a hard-coded architectural constraint. Inspired by the success of open-vocabulary detectors, we represent each target joint as a language-grounded query token. Specifically, we encode each bone name using a frozen CLIP text encoder and combine the semantic embedding with topological and directed kinematic descriptors. The three embeddings are concatenated and projected, then processed by a skeleton encoder with topology-aware attention. The resulting joint queries, augmented with character-specific residuals, enter a shared transformer decoder together with a global position token and image tokens from the current frame.

This formulation enjoys several key advantages. First, the decoder naturally handles rigs with arbitrary joint counts since the output is determined by the input query set. Second, the language-grounded nature of the queries ensures semantic consistency across diverse naming conventions, effectively bridging the gap between varying production standards. Empirically, we demonstrate that the combination of CLIP-based semantics and topological priors produces tokens that are discriminative within a single rig yet consistent across different characters. This property provides the essential foundation for robust, cross-rig motion synthesis without the need for manual correspondence authoring.

Building upon this semantic foundation, our framework further incorporates character-specific motion priors to achieve robust zero-shot generalization. While language-grounded queries effectively capture the universal semantic role of each bone, we enhance this representation during training by attaching learnable residuals that account for joint-specific range-of-motion and local kinematic coupling. As the model trains, these residuals naturally organize into a structured bone function space, where functionally equivalent joints seamlessly align. During inference on unseen rigs, we leverage this latent structure through a novel, non-parametric \emph{query-token retrieval} mechanism. For each joint in a novel topology, the system ranks training bones using a weighted combination of semantic and topological similarity, then interpolates their residuals with softmax weights to augment the novel rig's shared query. This strategy effectively closes the performance gap on unseen characters without requiring any retraining or incurring additional inference-time parameter costs.

We summarize our core contribution as follows. 
\begin{itemize}[leftmargin=*]
    \item We introduce a novel and real-time system, \emph{DragonMoCap}, capturing dragon motions from a video to diverse dragon-type characters, which is integrated as a UE add-on for practical animation workflows.
    \item We propose \emph{language-grounded joint query tokens} that fuse CLIP-based semantics with hierarchical topological descriptors, which ensures intra-rig discriminability and cross-rig similarity, establishing a robust foundation for cross-topology generalization.
    \item We introduce \emph{query-token retrieval}, a non-parametric mechanism that enables zero-shot generalization to unseen dragon rigs. This approach synthesizes high-quality query tokens for novel topologies without requiring any retraining or per-case optimization.
\end{itemize}

\begin{figure*}[t]
    \centering
    \includegraphics[width=\textwidth]{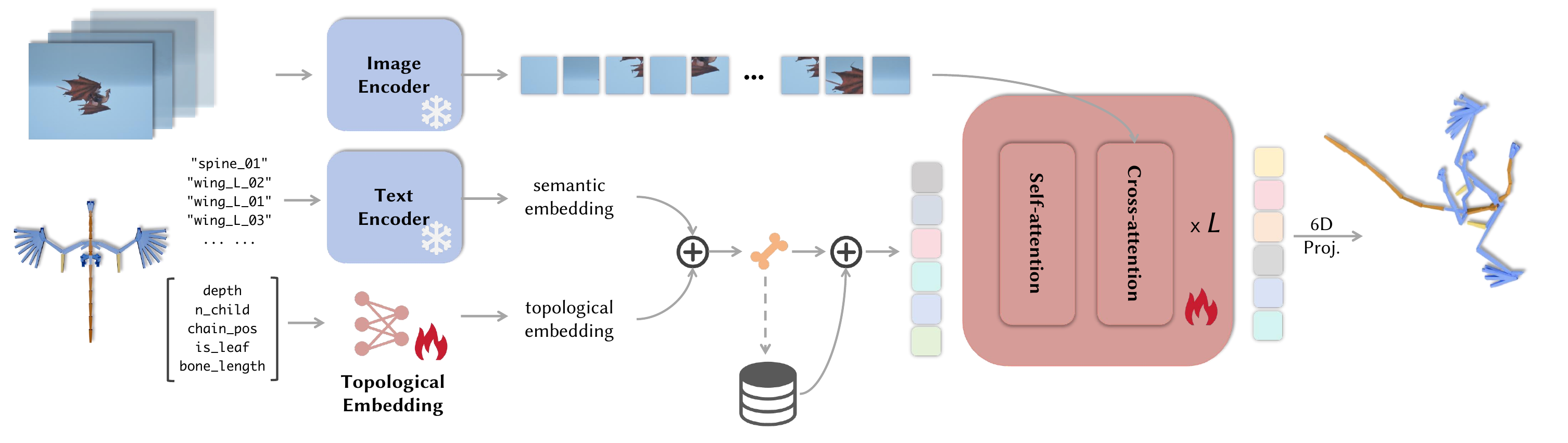}
    \caption{Method overview. \emph{DragonMoCap} is an image-based,
    per-frame predictor. For each input frame, a frozen image backbone
    produces image tokens. For each bone of the target rig, we build a
    \emph{language-grounded query token} by combining a CLIP bone-name
    embedding with a topological descriptor. The transformer decoder
    uses these query tokens to cross-attend to the image tokens and
    predicts per-joint rotations. For unseen rigs, missing per-bone
    residuals are synthesized by retrieving and interpolating the
    learned residuals of the nearest bones in a bone embedding bank.}
    \Description{A method diagram showing image tokens extracted by a
    frozen image backbone, language-grounded bone queries combining
    semantic and topological information, a transformer decoder for
    per-joint rotation prediction, and residual retrieval from a bone
    embedding bank for unseen dragon rigs.}
    \label{fig:method_overview}
\end{figure*}

\section{Related Work}

\subsection{Monocular Human Motion Capture}

Monocular human motion capture has been largely driven by parametric human body models such as SMPL~\cite{loper2015smpl} and SMPL-X~\cite{pavlakos2019expressive}. Early HMR-style methods directly regress human pose and shape from a single image~\cite{kanazawa2018hmr}, and later works improve robustness through optimization-in-the-loop training~\cite{kolotouros2019spin}, part-aware attention~\cite{kocabas2021pare}, mesh-alignment feedback~\cite{zhang2021pymaf}, hybrid inverse kinematics~\cite{li2021hybrik}, and full-frame location cues~\cite{li2022cliff}. Video-based methods further exploit temporal context for more stable reconstruction~\cite{kocabas2020vibe,choi2021tcmr}, while transformer-based systems such as HMR2.0/4DHumans improve image-based reconstruction and tracking~\cite{goel2023humans}. TokenHMR improves human body recovery through a tokenized pose representation~\cite{dwivedi2024tokenhmr}. Recent whole-body methods extend the output space from body-only SMPL to expressive SMPL-X recovery, including one-stage whole-body regression~\cite{osx} and large-scale expressive human pose estimation~\cite{cai2024smplerx,yin2025smplestx}. Despite these advances, existing HMR pipelines remain tied to human-specific priors and fixed SMPL/SMPL-X parameterizations, making them difficult to apply to dragon rigs with diverse joint counts, hierarchies, and naming conventions.
Synthetic datasets provide complementary supervision for human reconstruction.
SynBody~\cite{yang2023synbody} combines layered, clothed human models with
a scalable rendering pipeline to provide accurate 3D annotations for human
pose and shape estimation and neural rendering. Our UE5-rendered dragon
corpus similarly supplies paired images and motion labels, but targets
heterogeneous production rigs rather than a human parametric body model.

\subsection{Non-human Motion Capture, Synthesis, and Transfer}

For non-human subjects, several works reconstruct articulated animals or deformable objects from images and videos, including template-free articulated shape reconstruction~\cite{yang2021lasr}, neural animatable models from casual videos~\cite{yang2022banmo}, category-level deformable 3D object learning~\cite{wu2023dove}, and articulated animal reconstruction in the wild~\cite{wu2023magicpony}. These methods demonstrate the importance of moving beyond human templates, but they often focus on reconstructing shape, appearance, or category-level articulation rather than producing production-ready rotations for a user-specified rig.

In character animation, motion transfer and synthesis methods address heterogeneous skeletons from another direction. GenMM synthesizes motion from limited examples using generative motion matching~\cite{li2023genmm}, Motion2Motion transfers animation across different skeletal topologies with sparse correspondences~\cite{chen2025motion2motion}, and AnyTop generates motions for arbitrary skeletons using skeleton-conditioned diffusion~\cite{gat2025anytop}. More directly related to monocular arbitrary-skeleton capture, MoCapAnything formulates category-agnostic motion capture with a reference asset and factorized pose-to-rotation recovery~\cite{gong2026mocapanything}, while MoCapAnything V2 learns an end-to-end pose-to-rotation pipeline~\cite{gong2026mocapanythingv2}.
CAST~\cite{pang2026cast} is directly related to video-based capture for
arbitrary target skeletons. It combines learned local rotations with
predicted global bone directions and uses reliability-gated geometric
correction in parent-to-child order to reduce hierarchical rotation
errors. Our focus is language-grounded rig queries and retrieval of
learned bone residuals for unseen dragon rigs; our decoder directly
regresses rotations without a geometric correction stage.

\subsection{Automatic Rigging and Representations}

A parallel line of work studies how to make arbitrary 3D assets animatable by predicting skeletons and skinning weights. Classical and neural auto-rigging methods estimate animation skeletons and skinning for articulated characters~\cite{xu2020rignet}, incorporate motion cues for rigging~\cite{xu2022morig}, or learn skeletal articulations with neural blend shapes~\cite{li2021nbs}. Recent large-scale approaches further improve category generality through autoregressive or tokenized representations, including RigAnything~\cite{liu2025riganything}, UniRig~\cite{zhang2025unirig}, Anymate~\cite{deng2025anymate}, and SkinTokens~\cite{zhang2026skintokens}. These works are complementary to ours: they create or refine rigs, whereas \emph{DragonMoCap} assumes a user-provided rig and estimates its motion from video.

\section{Method}

\subsection{Overview and Problem Formulation}
\label{sec:overview}

\para{Setup.}
Given a monocular video $V=\{I_t\}_{t=1}^{T}$ with $T$ frames and a target
character $c$, \emph{DragonMoCap} recovers the motion of $c$ on its own rig.
Each character is associated with a single rig, so we use $c$ to index both.
The rig is a kinematic tree
$\mathcal{G}^{(c)}=(\mathcal{J}^{(c)},\mathcal{E}^{(c)})$ with joint set
$\mathcal{J}^{(c)}=\{1,\dots,N_c\}$, where $j=1$ denotes the root joint.
Each non-root joint $j$ has a parent $\pi(j)$, and the edge set
$\mathcal{E}^{(c)}=\{(\pi(j),j)\}_{j=2}^{N_c}$ encodes the parent--child
hierarchy. We denote by $\mathcal{A}(j)$ the set of ancestors of joint $j$,
by $\ell_{jk}$ the number of edges on the tree path between joints $j$ and
$k$ (so that $\ell_{\pi(j)j}=1$), and by $n_j$ the semantic bone name
attached to joint $j$.

For every frame $t\in\{1,\dots,T\}$, DragonMoCap predicts the root
translation $\hat{\mathbf{r}}_t\in\mathbb{R}^3$ in camera space and a
rotation $\hat{\boldsymbol{\theta}}_{t,j}\in\mathbb{R}^6$ for every joint
$j\in\mathcal{J}^{(c)}$.
The root rotation $\hat{\boldsymbol{\theta}}_{t,1}$ is the global
orientation of the character in camera space, while
$\hat{\boldsymbol{\theta}}_{t,j}$ for $j\geq 2$ is the local rotation of
joint $j$ relative to its parent $\pi(j)$.
All rotations use the continuous 6D representation.
For brevity, we omit the superscript $(c)$ whenever the rig is clear from
context.

\para{Rig-conditioned prediction.}
To support different dragon rigs with one network, we treat the rig as a
conditioning input. A shared decoder predicts each joint from a
rig-derived joint query, so its output cardinality follows the target rig
rather than a fixed joint vocabulary. Each joint query is the sum of two
parts: a \emph{shared query} $\mathbf{q}_j$, computed from the bone name,
skeletal topology, and hierarchical relations of joint $j$
(Sec.~\ref{sec:joint_query}), and a character-specific \emph{residual}
$\boldsymbol{\delta}_j$. Shared queries are computed in the same way for
any rig. Residuals are learned for training rigs and, for an unseen rig,
transferred from training rigs through non-parametric retrieval
(Sec.~\ref{sec:retrieval}).

\subsection{Per-Frame Image Encoder}
\label{sec:encoder}

Following the common paradigm of image-based human mesh
recovery~\cite{goel2023humans,dwivedi2024tokenhmr,yin2025smplestx},
we estimate motion by processing each video frame independently.
For each frame $I_t$, a frozen DINOv2
ViT-L/14~\cite{oquab2024dinov2} extracts spatially distributed visual
patch features $\mathbf{M}_t\in\mathbb{R}^{P\times d_{\mathrm{v}}}$,
where $P$ is the number of patch tokens and $d_{\mathrm{v}}$ is the
feature dimension.
These features serve as visual memory for the pose decoder
(Sec.~\ref{sec:decoder}), where the joint queries gather image evidence
through cross-attention.

\subsection{Language-Grounded Joint Queries}
\label{sec:joint_query}

\para{Query design principles.}
Our query design serves two goals.
First, \emph{intra-rig discriminability} allows the model to distinguish
joints with different roles, such as a finger and a leg, and to learn
their respective motion patterns.
Second, \emph{cross-rig similarity} allows joints in an unseen rig to
relate to similar joints encountered during training, providing cues for
predicting their motion.
To achieve both goals, we combine bone-name semantics for functional
similarity, topological and hierarchical-relation features for structural
context, and learned residuals for character-specific motion patterns.

\para{Semantic embedding.}
Bone names vary across rigs in naming conventions and descriptive detail.
Before text encoding, we use an LLM to rewrite each bone name $n_j$ into a
more consistent and descriptive form, making its semantic information more
explicit. A frozen CLIP text encoder~\cite{clip}, denoted
$E_{\mathrm{text}}$, then extracts the semantic descriptor
\begin{equation}
\mathbf{s}_j = E_{\mathrm{text}}\bigl(\mathrm{LLM}(n_j)\bigr),
\qquad
\mathbf{e}^{\mathrm{sem}}_j = \mathrm{MLP}_{\mathrm{sem}}(\mathbf{s}_j),
\label{eq:semantic}
\end{equation}
where $\mathbf{e}^{\mathrm{sem}}_j$ is the semantic embedding.
These features help the model learn shared motion patterns among joints
with similar semantic functions across different rigs. For an unseen
character, bone names provide cues about the likely motion roles of its
joints, supporting generalization to new skeletons.

\para{Topological and hierarchical-relation descriptors.}
Character rigs differ in skeletal topology. For each joint $j$, we
construct a topological descriptor $\mathbf{f}_j$ that collects its
hierarchy depth, child count, chain position, leaf status, subtree size,
and ordering cues, together with laterality and role indicators parsed
from the bone name. These features enable the model to learn reusable
patterns across similar skeletal structures.
To further describe how joint $j$ is situated in the kinematic tree, we
encode each of its directed ancestor and descendant relations, together
with the corresponding path length $\ell_{jk}$, using a shared entry-wise
encoder, and mean-pool the results into a relation descriptor
$\mathbf{f}^{\mathrm{rel}}_j$. Two MLPs then produce the topological and
relation embeddings:
\begin{equation}
\mathbf{e}^{\mathrm{topo}}_j = \mathrm{MLP}_{\mathrm{topo}}(\mathbf{f}_j),
\qquad
\mathbf{e}^{\mathrm{rel}}_j = \mathrm{MLP}_{\mathrm{rel}}(\mathbf{f}^{\mathrm{rel}}_j).
\label{eq:structural}
\end{equation}

\para{Shared query construction.}
For each joint $j$, we concatenate the three embeddings and fuse them with
a shared learned projection $\mathbf{W}_{\mathrm{in}}$ followed by GELU:
\begin{equation}
\mathbf{x}^{(0)}_j = \operatorname{GELU}\!\left(
\mathbf{W}_{\mathrm{in}}
\bigl[\mathbf{e}^{\mathrm{sem}}_j;\,
      \mathbf{e}^{\mathrm{topo}}_j;\,
      \mathbf{e}^{\mathrm{rel}}_j\bigr]\right).
\label{eq:query_fusion}
\end{equation}
A skeleton encoder then refines these features with local attention over
the kinematic tree, allowing each query to capture its structural context.
To make attention aware of the direction of skeletal relationships, we add
a hierarchy bias $\mathbf{H}\in\mathbb{R}^{N_c\times N_c}$, whose rows
index the attending joint $j$ and whose columns index the attended
joint $k$:
\begin{equation}
H_{jk}=
\begin{cases}
1, & j=k,\\
1/\ell_{jk}, & j\in\mathcal{A}(k)\ \ \text{($j$ is an ancestor of $k$)},\\
-0.5/\ell_{jk}, & k\in\mathcal{A}(j)\ \ \text{($k$ is an ancestor of $j$)},\\
0, & \text{otherwise}.
\end{cases}
\label{eq:hierarchy_bias}
\end{equation}
Let $\Omega(j)=\{k:\ell_{jk}\leq\rho\}$ denote the local neighborhood of
joint $j$ within $\rho$ hops. Each attention head $h$ learns a scale
$\gamma_h$ for the bias, and the attention weights are
\begin{equation}
A_{h,jk}=
\frac{\exp\bigl(Z_{h,jk}+\gamma_h H_{jk}\bigr)}
     {\sum_{k'\in\Omega(j)}\exp\bigl(Z_{h,jk'}+\gamma_h H_{jk'}\bigr)}
\quad\text{for } k\in\Omega(j),
\label{eq:skeleton_attention}
\end{equation}
and $A_{h,jk}=0$ otherwise. Here, $Z_{h,jk}$ is the scaled dot-product
score between the query of joint $j$ and the key of joint $k$ in head $h$.
After $L$ stacked attention and feed-forward layers, the skeleton encoder
outputs the shared queries $\mathbf{q}_j=\mathbf{x}^{(L)}_j\in\mathbb{R}^{d}$.
The same construction is used for training and unseen rigs.

\para{Learnable residuals.}
We further augment the shared queries with learnable tokens to increase
query expressiveness and capture motion patterns specific to each
character. Let $\mathcal{C}_{\mathrm{train}}$ denote the set of training
rigs. Each joint $j$ of a training rig $c\in\mathcal{C}_{\mathrm{train}}$
has a learnable residual $\boldsymbol{\delta}^{(c)}_j\in\mathbb{R}^{d}$,
initialized with small random values and added to its shared query:
\begin{equation}
\tilde{\mathbf{q}}^{(c)}_j=\mathbf{q}^{(c)}_j+\boldsymbol{\delta}^{(c)}_j.
\label{eq:adapted_query}
\end{equation}
We refer to $\tilde{\mathbf{q}}^{(c)}_j$ as the \emph{adapted query}.
After training, the residuals are collected into a bone bank for inference
on unseen rigs (Sec.~\ref{sec:retrieval}).

\subsection{Transformer Pose Decoder}
\label{sec:decoder}

For a rig with $N_c$ joints, the decoder operates on a query sequence of
$N_c+1$ tokens,
\begin{equation}
\mathbf{Q}=\bigl[\mathbf{q}_{\mathrm{tr}};\,
\tilde{\mathbf{q}}_1;\,\dots;\,\tilde{\mathbf{q}}_{N_c}\bigr]
\in\mathbb{R}^{(N_c+1)\times d},
\label{eq:decoder_input}
\end{equation}
consisting of a learned, rig-independent translation token
$\mathbf{q}_{\mathrm{tr}}\in\mathbb{R}^{d}$ and the $N_c$ adapted joint
queries, including that of the root joint. Each transformer layer applies
self-attention among the tokens of $\mathbf{Q}$, cross-attention from these
tokens to the image features $\mathbf{M}_t$, and a feed-forward block.
Let $\mathbf{h}_{t,\mathrm{tr}}$ and $\mathbf{h}_{t,j}$ denote the
final-layer outputs of the translation token and joint token $j$ for
frame $t$. Two prediction heads produce
\begin{equation}
\hat{\mathbf{r}}_t=\mathrm{MLP}_{r}(\mathbf{h}_{t,\mathrm{tr}}),
\qquad
\hat{\boldsymbol{\theta}}_{t,j}=\mathrm{MLP}_{\theta}(\mathbf{h}_{t,j}),
\quad j=1,\dots,N_c,
\label{eq:decoder_heads}
\end{equation}
where $\mathrm{MLP}_{\theta}$ is shared across all joints. Thus, the
root-joint token predicts the root orientation $\hat{\boldsymbol{\theta}}_{t,1}$,
and the remaining joint tokens predict local rotations.
Because the joint queries form a variable-length set and the rotation head
is shared, the decoder parameters are independent of joint count and
identity, allowing one decoder to operate on different dragon rigs without
changing its output head.

\subsection{Model Training}
\label{sec:training}

We train the projection MLPs, the skeleton encoder, the residuals
$\{\boldsymbol{\delta}^{(c)}_j\}$, the translation token, the decoder, and
the prediction heads with a per-frame pose loss. Both the image and text
encoders remain frozen. Given ground-truth joint rotations
$\boldsymbol{\theta}_{t,j}$, converted to the same 6D representation, and
ground-truth root translation $\mathbf{r}_t$, the loss for frame $t$ is
\begin{equation}
\mathcal{L}_{\mathrm{pose}}
=\frac{\lambda_{\theta}}{N_c}\sum_{j=1}^{N_c}
\bigl\|\hat{\boldsymbol{\theta}}_{t,j}-\boldsymbol{\theta}_{t,j}\bigr\|_1
+\lambda_r\bigl\|\hat{\mathbf{r}}_t-\mathbf{r}_t\bigr\|_1 ,
\label{eq:pose_loss}
\end{equation}
which is averaged over all frames in a mini-batch.
The rotation term includes the root joint and therefore supervises the
root orientation together with the remaining joint rotations.
The weights $\lambda_{\theta}$ and $\lambda_r$ balance rotation and
translation supervision. Normalizing the rotation term by $N_c$ prevents
rigs with more joints from dominating the batch objective.

\subsection{Model Inference on Unseen Rigs}
\label{sec:retrieval}

\para{Bone bank and candidate scoring.}
After training, we index every training bone by $b=(c,i)$ with
$c\in\mathcal{C}_{\mathrm{train}}$ and $i\in\mathcal{J}^{(c)}$, and denote
the index set by $\mathcal{I}_{\mathrm{train}}$. The bone bank stores the
semantic descriptor, the topological descriptor, and the learned residual
of each training bone,
\begin{equation}
\mathcal{B}=\bigl\{(\mathbf{s}_b,\,\mathbf{f}_b,\,\boldsymbol{\delta}_b)\bigr\}_{b\in\mathcal{I}_{\mathrm{train}}},
\qquad
(\mathbf{s}_b,\mathbf{f}_b,\boldsymbol{\delta}_b)
=(\mathbf{s}^{(c)}_i,\mathbf{f}^{(c)}_i,\boldsymbol{\delta}^{(c)}_i),
\label{eq:bone_bank}
\end{equation}
and remains fixed during inference. Here, $\mathbf{s}$ and $\mathbf{f}$
are semantic embedding and the descriptors defined in Eq.~\eqref{eq:semantic} and
Sec.~\ref{sec:joint_query}.
For an unseen rig $u$, we first compute its shared queries
$\mathbf{q}^{(u)}_j$ with the same semantic, topological, and
hierarchical-relation encoding used for training rigs.
For each target joint $j\in\mathcal{J}^{(u)}$, every bank entry $b$
receives a weighted semantic and topological similarity score
\begin{equation}
a_{jb}=\alpha\,\operatorname{sim}\bigl(\mathbf{s}^{(u)}_j,\mathbf{s}_b\bigr)
+\beta\,\operatorname{sim}\bigl(\mathbf{f}^{(u)}_j,\mathbf{f}_b\bigr),
\label{eq:retrieval_score}
\end{equation}
where $\operatorname{sim}$ denotes cosine similarity, and $\alpha$ and
$\beta$ weight the semantic and topological terms.
Semantic similarity identifies bones with related meanings, while
topological similarity helps distinguish similarly named bones that occupy
different positions in the skeleton hierarchy.
Let $\mathcal{N}_m(j)\subset\mathcal{I}_{\mathrm{train}}$ contain the $m$
highest-scoring bank entries for joint $j$.

\para{Residual interpolation.}
We transfer the selected residuals with temperature-scaled softmax weights:
\begin{equation}
\bar{\boldsymbol{\delta}}^{(u)}_j=
\sum_{b\in\mathcal{N}_m(j)}
\frac{\exp(a_{jb}/\tau)}
     {\sum_{b'\in\mathcal{N}_m(j)}\exp(a_{jb'}/\tau)}\,
\boldsymbol{\delta}_b ,
\label{eq:interpolation}
\end{equation}
where the temperature $\tau$ controls how strongly the transfer favors the
highest-scoring entries.
The unseen rig retains its own shared queries and receives only the
interpolated residuals from training rigs. Mirroring
Eq.~\eqref{eq:adapted_query}, its adapted queries are
\begin{equation}
\tilde{\mathbf{q}}^{(u)}_j=\mathbf{q}^{(u)}_j+\bar{\boldsymbol{\delta}}^{(u)}_j ,
\label{eq:unseen_query}
\end{equation}
which are fed to the decoder in place of the training-rig queries in
Eq.~\eqref{eq:decoder_input}.
The procedure requires no retraining or per-rig optimization, only
candidate scoring and interpolation over the fixed bone bank.
Sec.~\ref{sec:residual_analysis} examines the semantic structure of the
learned residuals that supports this transfer.

\section{Experiments}

\subsection{Implementation Details}
\label{sec:implementation_details}

\para{Model implementation.}
We use a frozen DINOv2 ViT-L/14~\citep{oquab2024dinov2} as the per-frame
image encoder and a frozen CLIP ViT-L/14~\citep{clip} text tower for
bone-name grounding. The transformer decoder has $8$ layers, a hidden
dimension of $512$, and $8$ attention heads, and is applied independently
to each frame. All rig-dependent quantities, including CLIP bone-name
embeddings, topological and hierarchical-relation descriptors, and shared
queries, are computed once per rig and cached. Excluding the CLIP text
tower, whose outputs are cached and never evaluated at inference time,
the model comprises approximately $420\mathrm{M}$ parameters, of which
$115\mathrm{M}$ are trainable. We train on $4\times$ NVIDIA A100 GPUs
with an effective batch size of $512$ and input resolution
$308\times308$, using AdamW~\citep{adamw} with an initial learning rate
of $2\times10^{-4}$ for $200$ epochs.

\para{Appearance augmentation.}
To reduce overfitting to the limited visual diversity of UE5 dragon
assets, we apply Gemini-based appearance edits (skin texture, color
palette, and background) to a small set of pose-clustered frames per
rig, preserving the original pose and rotation labels.

\subsection{Experimental Setup}
\label{sec:exp_setup}

\para{Data.}
We use an internal dataset of professionally animated dragon clips
rendered in Unreal Engine~5, with ground-truth joint rotations and root
transforms exported directly from the engine. It contains $8$ training
rigs and $7$ held-out rigs with $20$--$60$ joints each, covering three
topology classes in roughly balanced proportions: winged quadrupeds,
wingless quadrupeds, and winged bipeds (wyvern-style). Each training rig
provides $40$K--$120$K animator-authored frames, which we augment with
SLERP-interpolated poses to about $150$K frames per rig ($\sim1.2$M in
total). All quantitative evaluation is conducted on UE5-rendered held-out
data; AI-generated inputs without ground truth are evaluated through a
user study (Sec.~\ref{sec:user_study}).

\para{Evaluation protocol.}
We evaluate in two settings: (1) \textit{seen rigs}, where the $8$
training rigs are evaluated on held-out clips with unseen motions; and
(2) \textit{unseen rigs}, where the $7$ held-out rigs share topology
classes with the training rigs but differ in rig definitions, bone
counts, and naming conventions, serving as the main cross-rig
generalization setting. We report \textit{MPJPE} (cm), the world-space
mean per-joint position error, and \textit{PA-MPJPE} (cm), its
Procrustes-aligned variant, which factors out a per-frame similarity
transform and focuses on articulation quality. Each rig is normalized to
a $1.6$--$2$\,m bounding size to control for scale differences.

\para{Baselines and variants.}
Since prior work on dragon motion reconstruction is limited, we adopt
two representative SMPL-X-based human motion reconstruction methods,
\textit{SMPLest-X}~\cite{yin2025smplestx} and \textit{OSX}~\cite{osx},
and retrain both on our dragon training data. For fairness, we replace
their image encoders with ours and adjust the remaining capacity to
match our number of trainable parameters. Their output heads are defined
over the joints of the training rigs, so they can only be evaluated on
seen rigs. \textit{MoCapAnything (V1 \&
V2)}~\citep{gong2026mocapanything,gong2026mocapanythingv2} also targets
non-humanoid motion capture, but official code and weights were not
available at the time of our experiments.
We further compare three query-design variants that differ from the full
model only in how joint queries are constructed during training:
\textit{learned embedding} uses a free learnable embedding per joint,
while \textit{CLIP only} and \textit{topology only} build the shared
query from the semantic or the topological and hierarchical-relation
embeddings alone.
On unseen rigs, all variants use the same residual retrieval
(Sec.~\ref{sec:retrieval}), so their differences reflect query
construction rather than the transfer mechanism.

\subsection{Main Results}

\begin{table}[t]
\centering
\caption{Quantitative comparison and query-design ablation on seen and
unseen dragon rigs (MPJPE / PA-MPJPE in cm, lower is better). All
methods are trained on our dragon data. External baselines (``---'')
cannot be evaluated on unseen rigs because their output heads are
defined over the training-rig joints. All variants use the same residual
retrieval on unseen rigs.}
\Description{A table comparing external baselines and query designs using MPJPE and PA-MPJPE on seen and unseen dragon rigs.}
\label{tab:main}
\footnotesize
\setlength{\tabcolsep}{2pt}
\begin{tabular}{lcccc}
\toprule
 & \multicolumn{2}{c}{Seen rigs} & \multicolumn{2}{c}{Unseen rigs} \\
\cmidrule(lr){2-3} \cmidrule(lr){4-5}
Method & MPJPE$\downarrow$ & PA-MPJPE$\downarrow$ & MPJPE$\downarrow$ & PA-MPJPE$\downarrow$ \\
\midrule
SMPLest-X~\citeyearpar{yin2025smplestx}
  & 10.57 & 6.53 & --- & --- \\
OSX~\citeyearpar{osx}
  & 11.20 & 7.12 & --- & --- \\
\midrule
Learned embedding
  & 10.24 & 6.58 & 14.62 & 9.32 \\
CLIP only
  & 10.51 & 7.20 & 15.32 & 9.59 \\
Topo only
  & 10.84 & 6.67 & 14.82 & 9.63 \\
\midrule
\textbf{Ours}
  & \textbf{9.46} & \textbf{6.01} & \textbf{12.11} & \textbf{8.74} \\
\bottomrule
\end{tabular}
\end{table}

As shown in Table~\ref{tab:main}, \emph{DragonMoCap} achieves the best
performance in both settings. On seen rigs, it reduces MPJPE from
$10.57$ to $9.46$ ($-10.5\%$) relative to the strongest retrained
baseline, SMPLest-X, showing that a unified, rig-conditioned query
representation does not sacrifice accuracy on familiar rigs. On unseen
rigs, residual retrieval transfers learned residuals without any
training data for the target rig, and our model reaches $12.11$ MPJPE
and $8.74$ PA-MPJPE, outperforming all query-design variants
($14.62$--$15.32$ MPJPE). Since all variants share the same retrieval
procedure, this gain is attributable to the combined semantic and
structural query design.

\para{Empirical validation of query design.}
We verify that combining CLIP and topological features satisfies both
properties in \cref{sec:joint_query} through a cross-rig bone retrieval
experiment: given a bone from one dragon, we retrieve its semantically
equivalent bone in another dragon by the cosine similarity between their
queries. The experiment covers $8$ dragon rigs, with $1{,}252$ queries in
total and an average of $47.9$ candidate bones. Queries built from CLIP
features alone show strong cross-rig similarity for corresponding bones,
but often confuse bones sharing the same prefix at different depths,
failing the intra-rig discriminability property. Topological features
alone resolve such structural ambiguities but lack the semantic
information needed to distinguish different parts. Their combination
yields the most reliable retrieval, confirming that CLIP captures
\emph{what} a bone is, while topology captures \emph{where} it is
located.

\para{Effect of query design on motion capture.}
The learned-embedding, CLIP-only, and topology-only variants in
Table~\ref{tab:main} share the same decoder, training objective, and
residual retrieval, so their differences isolate the effect of query
construction. Neither CLIP-only nor topology-only queries outperform
learned embeddings, while the full model, which combines both
descriptors, achieves the lowest MPJPE and PA-MPJPE in both settings. On
unseen rigs, it reduces MPJPE from $14.62$ for learned embeddings to
$12.11$, supporting the cross-rig generalization of the combined design.

\begin{figure*}[h]
    \centering
    \includegraphics[width=\linewidth]{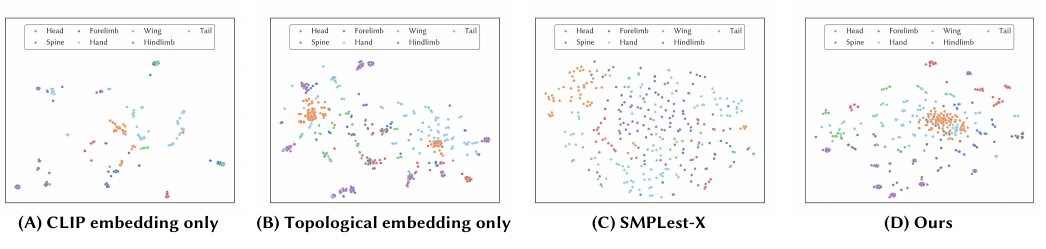}
    \vspace{-2.5em}
    \caption{\textbf{t-SNE visualization of query embeddings.} 
    In contrast to baselines that suffer from a lack of global coherence or intermingled features, our integrated approach effectively resolves ambiguities between anatomically similar parts to achieve superior, semantic-aware clustering.}
    \label{fig:tsne_analysis}
\end{figure*}

To qualitatively analyze these effects, we provide t-SNE visualizations
of the bone embeddings in \cref{fig:tsne_analysis}. As illustrated in
\cref{fig:tsne_analysis}, the representations in (A), (B), and (C) are
significantly less discriminative. Specifically, using only CLIP
embeddings (A) leads to fragmented and sparse clusters, while relying
solely on topological embeddings (B) fails to distinguish between
anatomically similar parts, such as the ``\texttt{Forelimb}'' (blue) and
``\texttt{Hindlimb}'' (purple), which appear heavily overlapped. Baseline
methods (C) like SMPLest-X (learned embedding only) exhibit a highly
chaotic distribution where different semantic categories are
intermingled. In contrast, our full design (D) achieves the most
cohesive and distinct clustering, effectively separating even
functionally similar limbs into well-defined groups. This visual
evidence confirms that the synergy of CLIP-based semantic priors and
structural topology is essential for learning a robust, rig-agnostic
latent space.

\subsection{Qualitative Results}
\label{sec:qualitative_real}

\begin{figure*}[t]
    \centering
    \includegraphics[width=\linewidth]{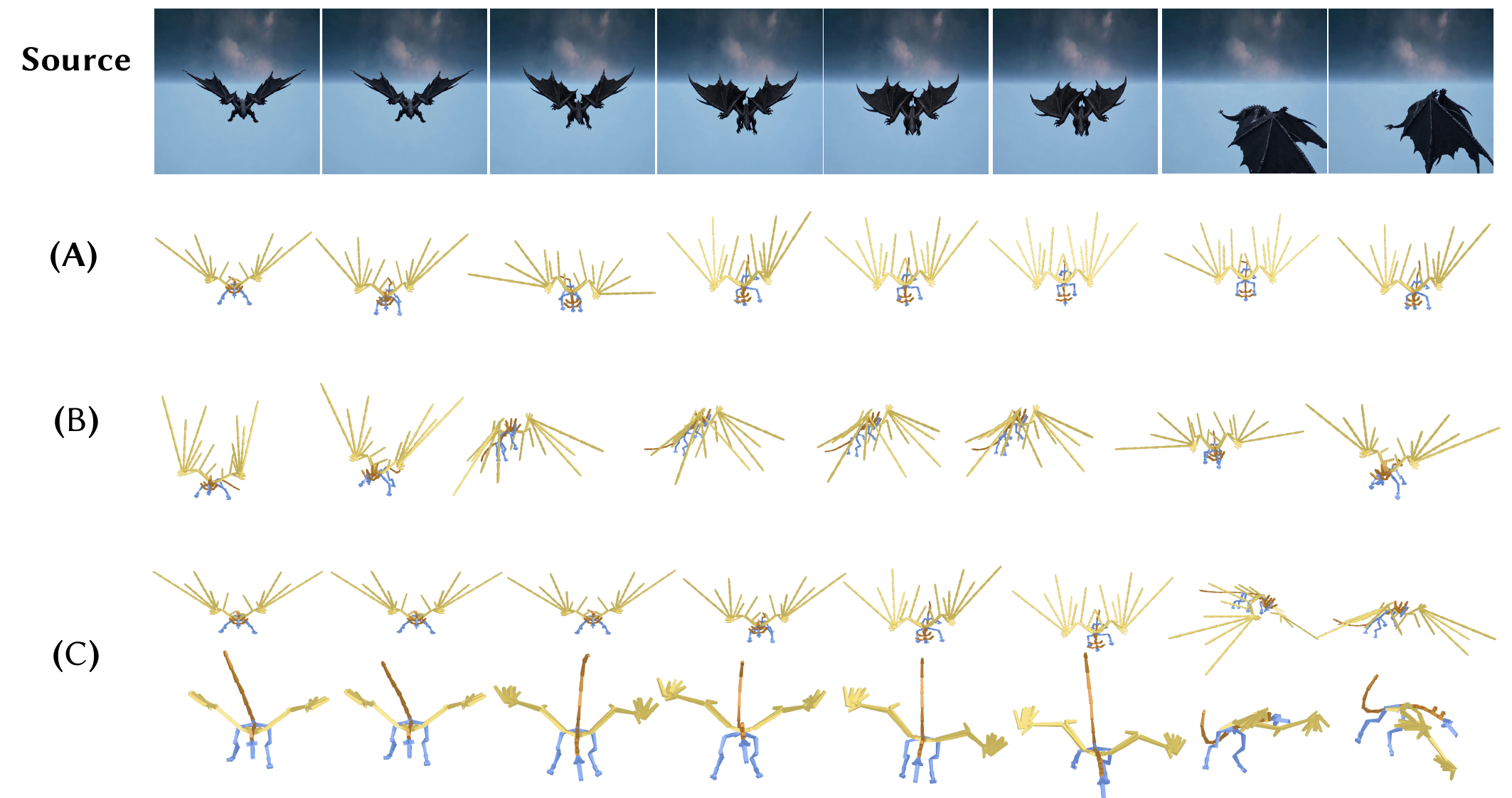}
    \vspace{-2em}
    \caption{\textbf{Qualitative results.} We take a video as input, and the captured dragon is performed by SMPLest-X (A), OSX (B), and Ours (C). As baselines cannot be generalized to novel dragons, we only provide results on one target rig. Moreover, we provide another capture result for unseen rig at the bottom of (C).}
    \label{fig:qualitative}
\end{figure*}

\begin{figure*}[t]
    \centering
    \includegraphics[width=\linewidth]{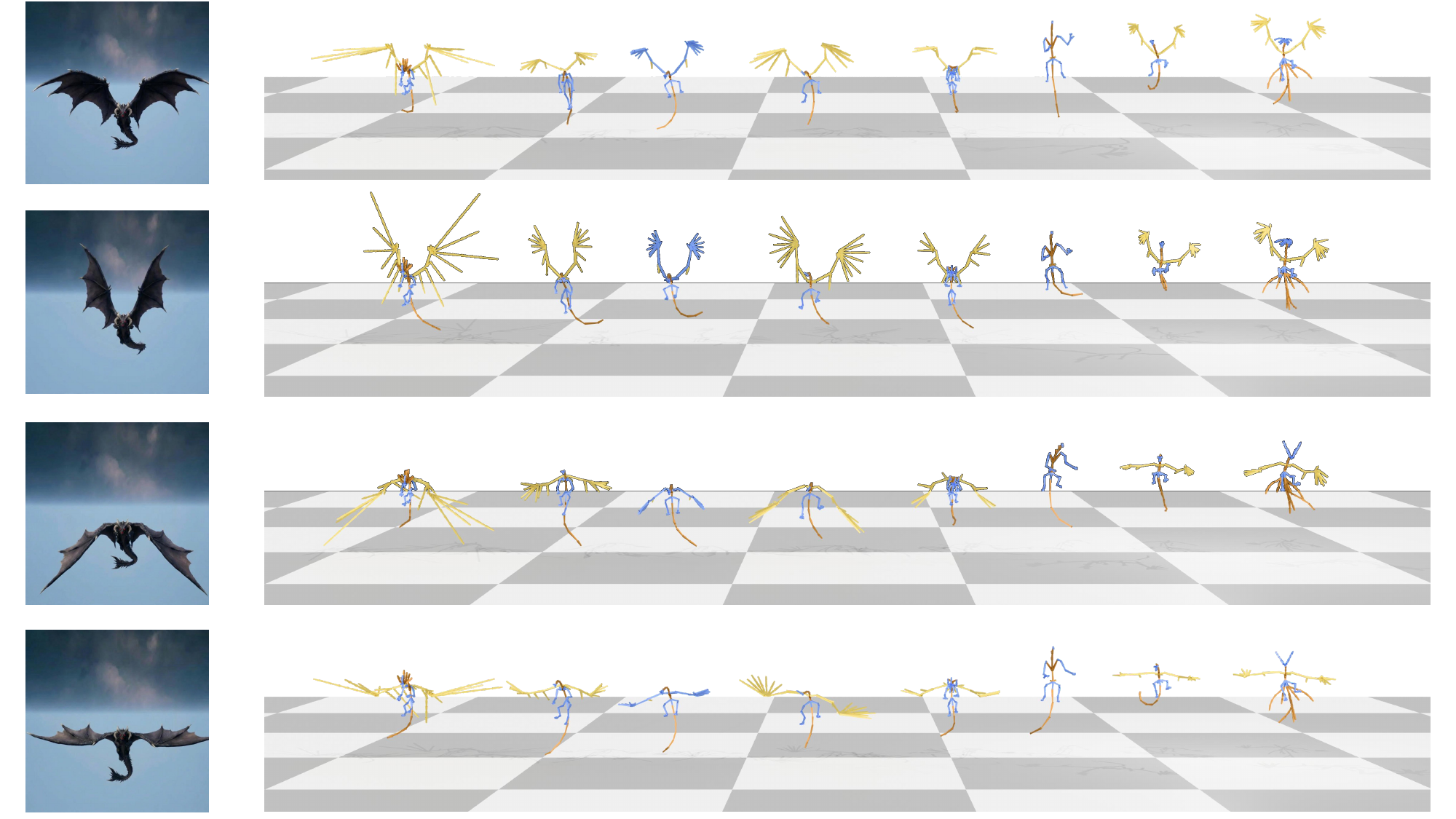}
    \vspace{-2em}
    \caption{One-video-to-many-rigs. A single input dragon video (left) simultaneously drives 8 dragon rigs of different topologies (right), producing coordinated yet rig-appropriate motion. Especially, the left 4 rigs are seen during training, and the right 4 rigs are unseen for our model.}
    \vspace{-1em}
    \label{fig:one2many}
\end{figure*}

Figure~\ref{fig:qualitative} compares \emph{DragonMoCap} with the
competing baselines on two seen-rig cases, a winged quadruped with an
articulated tail and a winged biped (wyvern) with wing fans, and
additionally shows our motion capture results on an unseen rig with
a distinct joint layout, to which the baselines cannot be applied.
Figure~\ref{fig:one2many} illustrates a capability that follows directly
from rig-conditioned decoding: simultaneously driving several dragon
rigs of different topologies from a single input video, with each output
respecting its own rig's structure and without training data for
the target rigs.

\subsection{User Study}
\label{sec:user_study}

\begin{table}[t]
\centering
\caption{User study on AI-generated inputs ($40$ participants $\times$
$30$ clips $=1{,}200$ paired trials per criterion). Left: mean $5$-point
Likert score. Right: win rate of ours over SMPLest-X, with ties counted
as half a win.}
\Description{A user study table comparing motion fidelity, anatomical plausibility, and visual quality between SMPLest-X and DragonMoCap.}
\label{tab:user_study}
\footnotesize
\setlength{\tabcolsep}{2pt}
\begin{tabular}{lccc|c}
\toprule
 & \multicolumn{2}{c}{Mean score$\uparrow$} & & Win rate$\uparrow$ \\
\cmidrule(lr){2-3}
Criterion & SMPLest-X & \textbf{Ours} & $p$ & \shortstack{Ours vs.\\ SMPLest-X} \\
\midrule
Motion fidelity
  & 3.7 & \textbf{4.4} & $<\!0.001$ & \textbf{72.5\%} \\
Anatomical plausibility
  & 3.2 & \textbf{4.5} & $<\!0.001$ & \textbf{83.6\%} \\
Overall visual quality
  & 4.1 & \textbf{4.6} & $<\!0.001$ & \textbf{74.8\%} \\
\bottomrule
\end{tabular}
\end{table}

Since AI-generated dragon videos have no ground-truth motion, we conduct
a user study. Forty participants compared \emph{DragonMoCap} with the
retrained SMPLest-X on $30$ AI-generated dragon clips. The two outputs
were shown side by side in random order without method labels and rated
for motion fidelity, anatomical plausibility, and overall visual quality
on a $5$-point Likert scale; paired differences are assessed with the
Wilcoxon signed-rank test. As shown in Table~\ref{tab:user_study},
participants consistently preferred \emph{DragonMoCap}, especially for
anatomical plausibility ($+1.3$ Likert, $83.6\%$ win rate, $p<0.001$),
indicating more natural dragon motion even when both methods are trained
on the same data.

\subsection{Residual Space Analysis}
\label{sec:residual_analysis}

Residual retrieval assumes that residuals learned for functionally
equivalent bones in different training rigs are similar. To verify this,
we group all learned residuals by semantic bone category (spine, limb,
wing, tail, head, extremity) and compute pairwise cosine similarity
across rigs. Within a category, residuals from different rigs reach
$>$$0.85$ similarity on average for spine, limb, and tail bones, whereas
similarity across categories drops below $0.2$. Although each residual
is learned for a specific joint of a specific character, residuals thus
converge to motion priors organized by functional role rather than by
character identity: the shared query specifies which joint is predicted,
and the residual adds the motion characteristics of its role. This
shared structure is what makes retrieval-based transfer possible.

\para{Retrieval quality.}
In a leave-one-rig-out experiment, where each training rig is removed
from the bone bank and its joints are retrieved from the remaining rigs,
retrieval achieves $78.3\%$ top-1 and $94.6\%$ top-5 semantic-role
match, supporting residual transfer between functionally related bones.

\subsection{Explainability}

To examine whether the model attends to semantically meaningful regions,
we visualize the cross-attention maps between bone queries and input
frames (\cref{fig:cross_attention}). Queries for anatomical parts such
as \texttt{``Head''} consistently activate the dragon's head across
different poses and rigs (rows A and B). The model also distinguishes
symmetric components, such as \texttt{``Wrist\_L''} and
\texttt{``Wrist\_R''}, attending precisely to the respective wing
joints. This indicates that CLIP-based semantic grounding and
topological priors allow the decoder to ``look'' at the correct
locations when estimating the motion of each bone, rather than relying
on purely statistical correlations.

\subsection{Runtime Efficiency}
\label{sec:runtime}

On a single NVIDIA L20 GPU, the full per-frame pipeline (DINOv2 encoder
and rig-conditioned decoder) runs at $\sim$30~FPS, sufficient for
real-time preview and efficient offline processing of long clips.
Rig-level quantities are computed once per rig and cached, adding no
per-frame cost. When driving multiple rigs from one video, the image
encoder is shared and only the lightweight decoder is re-run per rig,
so the marginal cost of each additional rig is minor. This efficiency
enables integration into mainstream animation workflows such as UE5
(\cref{sec:interface}).

\section{Application}

\begin{figure*}[!t]
    \centering
    \begin{minipage}[t]{0.48\linewidth}
        \centering
        \includegraphics[width=\linewidth]{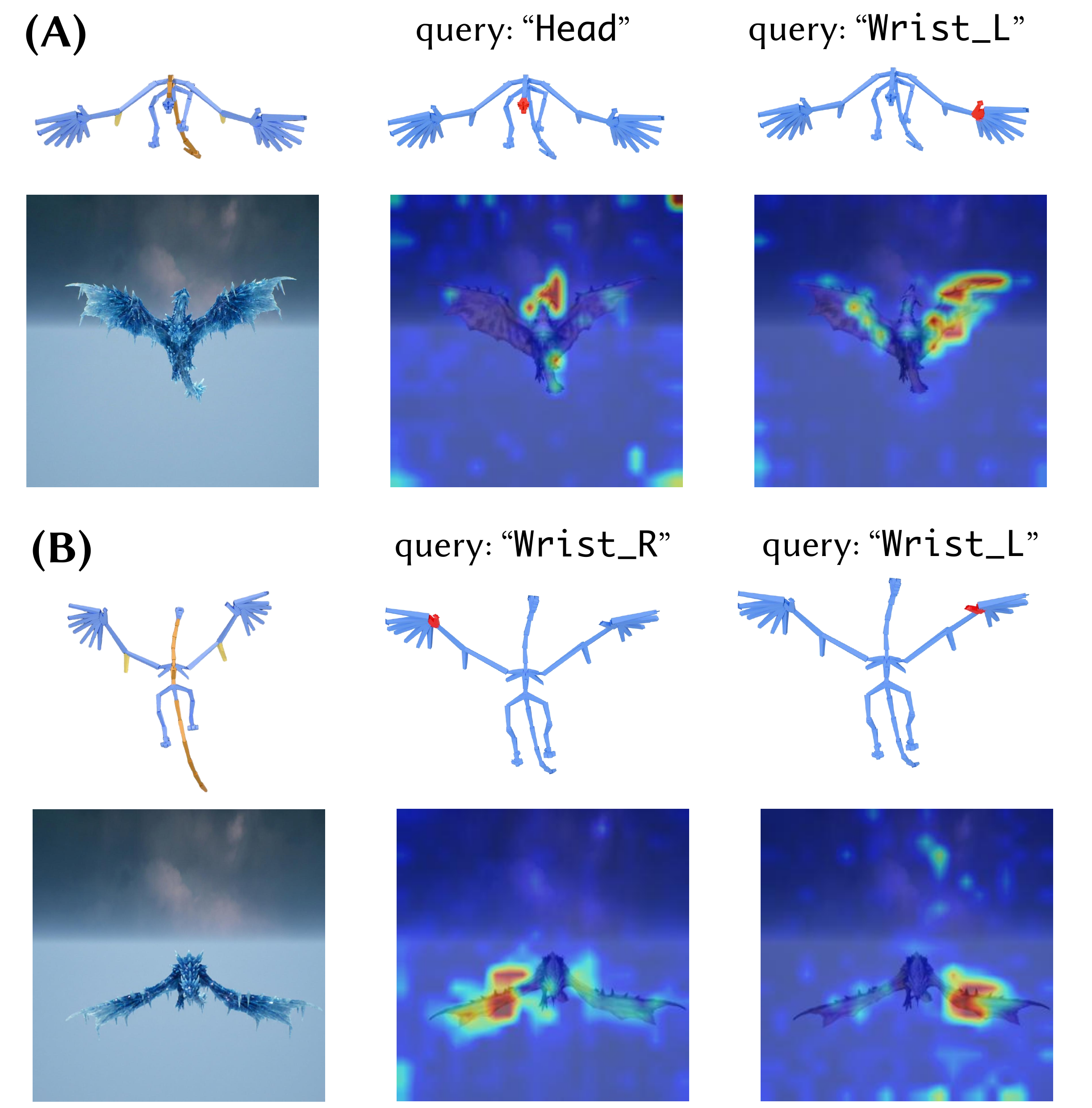}
        \vspace{-2em}
        \caption{\textbf{Cross-attention visualization.} 
        We visualize how different bone queries attend to image features. The query bones are denoted in red. In both (A) and (B), the model accurately localizes the corresponding anatomical regions (\textit{e.g.}, ``\texttt{Head}'', ``\texttt{Wrist}''). The sharp activation on symmetrical parts like ``\texttt{Wrist\_L}'' and ``\texttt{Wrist\_R}'' highlights the strong discriminative power of our query design.}
        \label{fig:cross_attention}

        \includegraphics[width=\linewidth]{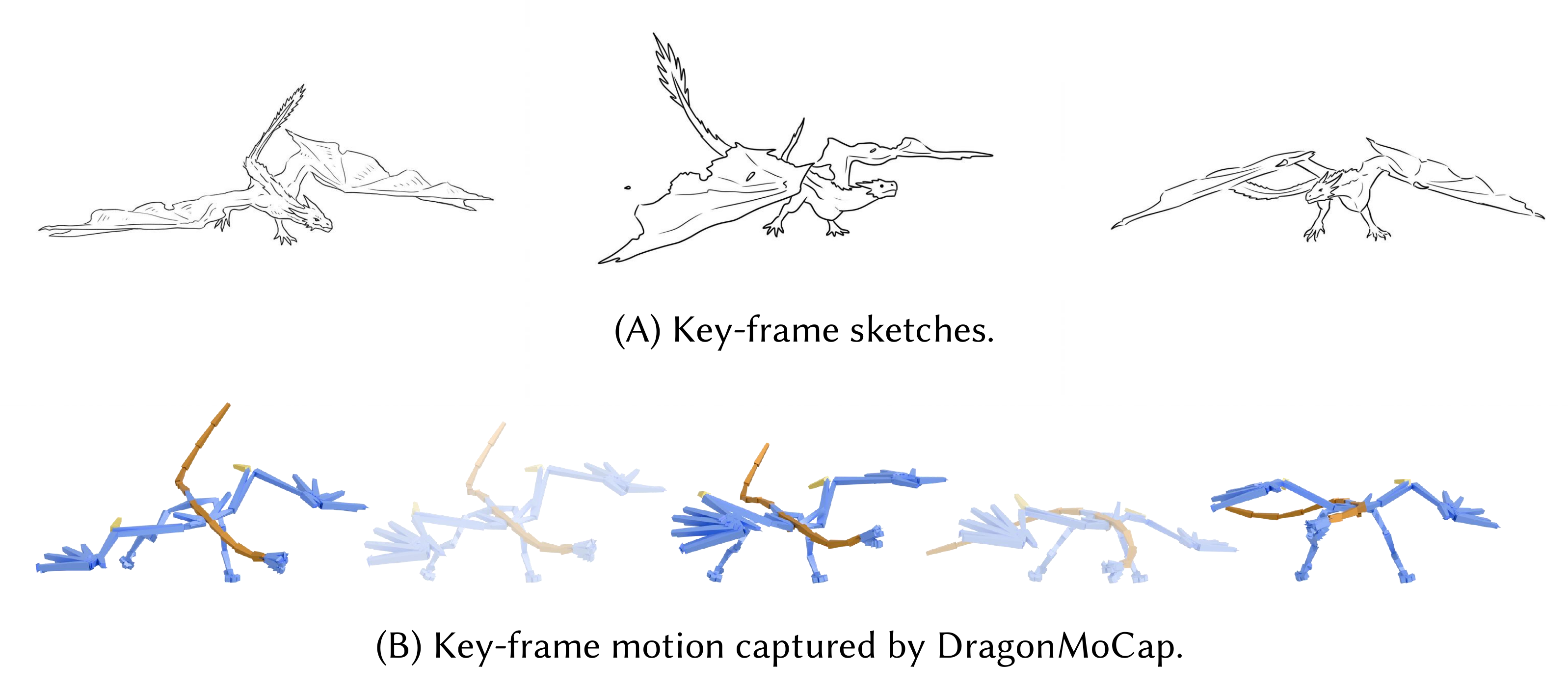}        \vspace{-2em}
        \caption{\textbf{Sketch-to-motion synthesis.} (A) Input sparse key-frame sketches provided by an artist. (B) The synthesized motion sequence on the target rig. Our \emph{DragonMoCap} captures skeletal poses as constraints (solid colors), then generates dense animation via interpolation (semi-transparent).}
        \label{fig:sketch_to_motion}        \vspace{-5em}
    \end{minipage}
    \hfill \begin{minipage}[t]{0.48\linewidth}
        \centering
        \includegraphics[width=\linewidth]{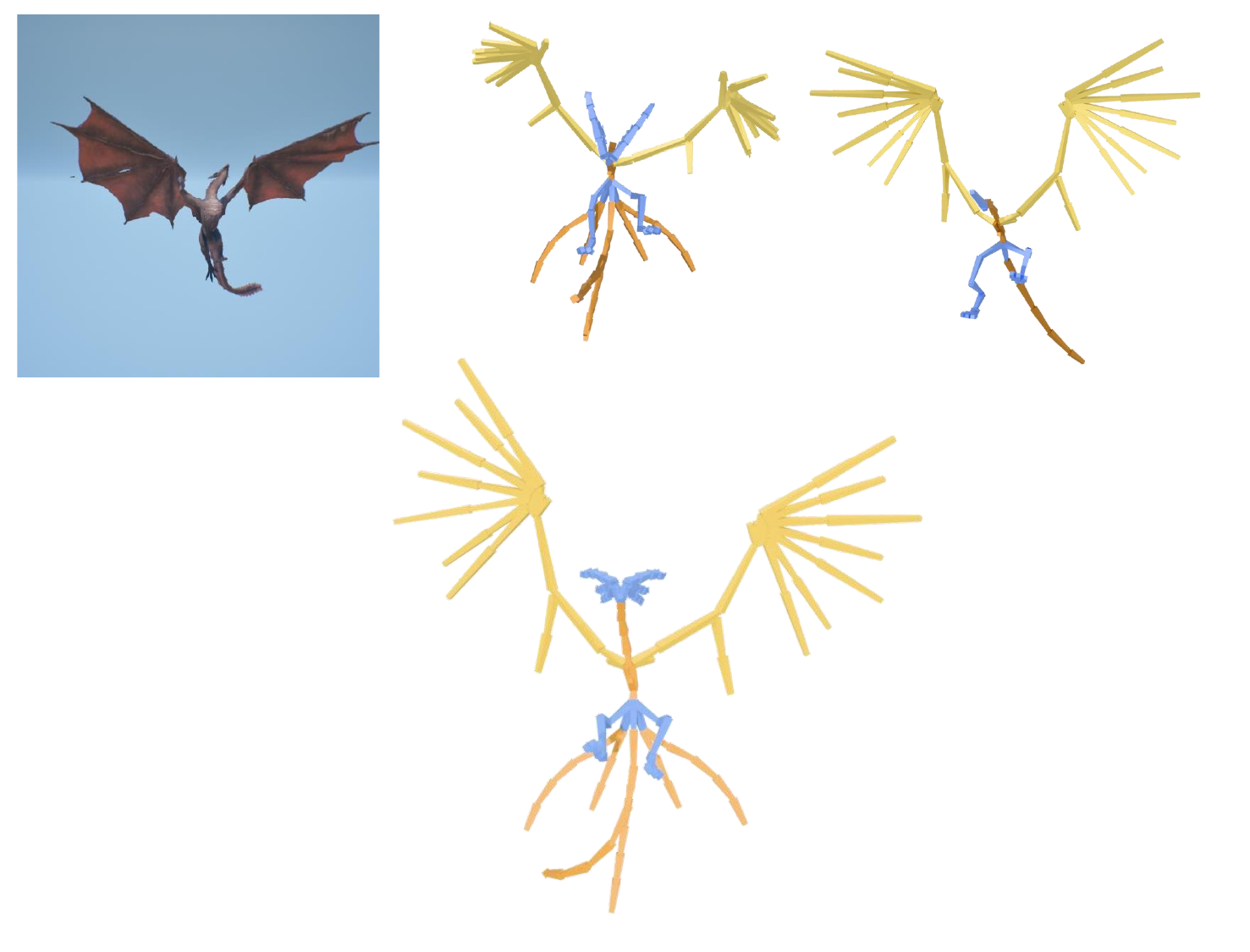}        \vspace{-2em}

        \caption{\textbf{Generalization to novel skeletons.} Given a single input dragon image, our method synthesizes plausible results for different target skeletons. The first row shows the input, the result with skeleton A, and the result with skeleton B. The second row shows the result on a novel hybrid skeleton combining the body of dragon A with the wings of dragon B. Our method remains coherent on this unseen skeleton, demonstrating strong generalization.}
        \label{fig:motion_reassembly}

        \includegraphics[width=\linewidth]{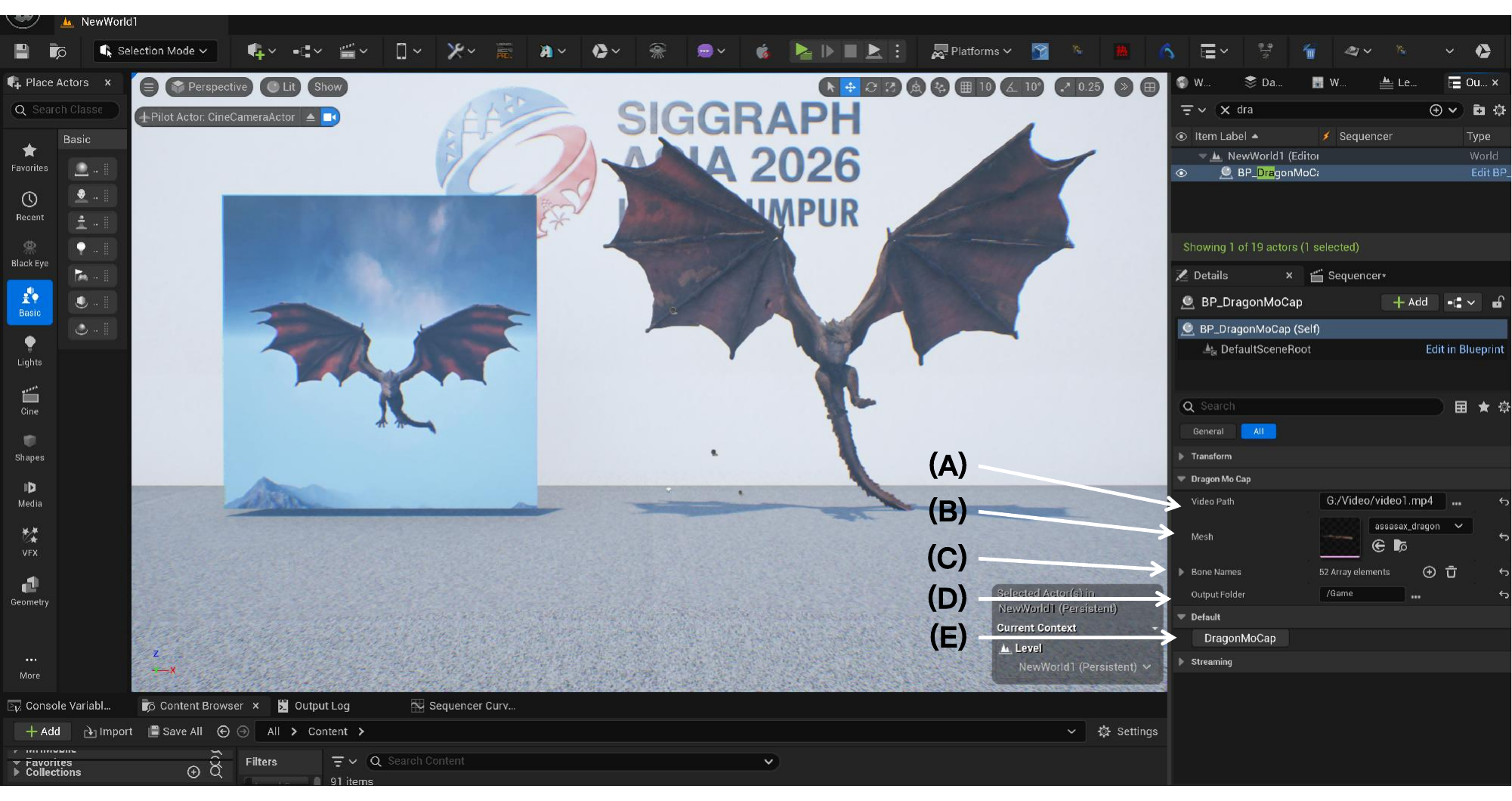}        \vspace{-2em}

        \caption{\textbf{UE-5 Plugin for \emph{DragonMoCap}.} 
        Our dedicated plugin provides a streamlined interface: (A) video selection, (B) mesh assignment, (C) bone filtering, and (D) output configuration. (E) initiates the system, enabling synchronized comparison between source (F) and result (G).}        \vspace{-2em}

        \label{fig:addon}
    \end{minipage}
            \vspace{-2em}
\end{figure*}

\para{Synthesizing motion from key-frame sketches.}
Beyond video-based capture, our framework enables a streamlined sketch-to-motion workflow. As illustrated in Fig.~\ref{fig:sketch_to_motion}, an artist can provide a sparse set of expressive key-frame sketches (A). \emph{DragonMoCap} treats each sketch as an input to recover its corresponding 3D pose, which then serves as a key-frame constraint for a motion interpolation algorithm (e.g., GenMM~\cite{li2023genmm}). By synthesizing intermediate frames between these recovered poses (shown as solid \textit{vs.} semi-transparent rigs in B), the system automatically generates a dense and temporally coherent animation sequence on the target rig. This interface allows artists to produce complex dragon animations by specifying only a few critical poses.

\para{Motion reassembly from one video.}
Our model generalizes well to unseen reassembled skeletons. Specifically, we construct hybrid rigs by combining parts from different characters, e.g., attaching the wings of dragon A to the body of dragon B. Although such combinations are not seen during training, our method still produces plausible and coherent motion results, as shown in \cref{fig:motion_reassembly}. This indicates that the model captures transferable part-level motion priors and can handle novel bone compositions.

\section{Animation User Interface}
\label{sec:interface}

To facilitate the seamless integration of our model into professional production pipelines, we developed a dedicated Unreal Engine 5 (UE-5) plugin leveraging UE-5 editor framework and reflection system. 
As illustrated in \cref{fig:addon}, the streamlined GUI panel serves as the primary interface for motion capture configuration. 
Users first select the target video (A) via an integrated file picker, which is automatically rendered by an in-scene viewer (F). 
After assigning the dragon mesh (B), the user specifies a subset of bones (C) for capture, effectively filtering out auxiliary gameplay bones that do not require animation. 
Once the output directory (D) is defined, clicking the \textit{DragonMoCap} button (E) initiates the backend processing. 
The system tracks the progress in real-time and subsequently compiles the returned motion data into a native UE-5 animation asset. 
To support efficient evaluation, the plugin features synchronized playback, allowing users to perform a rigorous side-by-side comparison between the original footage (F) and the resulting character motion (G) within the engine 3D viewport.

\section{Conclusion}

We presented \emph{DragonMoCap}, to our knowledge the first
video-based motion-capture system for dragon-type characters.
By treating the target rig as an input rather than an architectural
constant, and specifying it through \emph{language-grounded query
tokens} that fuse semantic, topological, and kinematic features through a
shared skeleton encoder, a single decoder operates on heterogeneous dragon rigs; a non-parametric
\emph{query-token retrieval} further extends this to rigs never seen
during training. This design makes the system practical for production
settings, where character rigs often differ substantially across assets
and are rarely standardized. On our dataset of $8$ training and $7$
held-out dragon rigs, \emph{DragonMoCap} outperforms HMR-style and
learned-embedding baselines in every setting and remains usable on
AI-generated inputs. These results suggest that language- and
structure-conditioned motion capture is a promising direction for
animating non-human characters with diverse skeletal layouts. The main
remaining bottleneck is the image encoder's appearance coverage, which
we mitigate with generative appearance augmentation and a
first/last-frame-conditioned production pipeline; scaling this further,
together with extending beyond dragons, is our primary future direction.

\clearpage

\clearpage

\begin{thebibliography}{36}

\ifx \showCODEN    \undefined \def \showCODEN     #1{\unskip}     \fi
\ifx \showDOI      \undefined \def \showDOI       #1{#1}\fi
\ifx \showISBNx    \undefined \def \showISBNx     #1{\unskip}     \fi
\ifx \showISBNxiii \undefined \def \showISBNxiii  #1{\unskip}     \fi
\ifx \showISSN     \undefined \def \showISSN      #1{\unskip}     \fi
\ifx \showLCCN     \undefined \def \showLCCN      #1{\unskip}     \fi
\ifx \shownote     \undefined \def \shownote      #1{#1}          \fi
\ifx \showarticletitle \undefined \def \showarticletitle #1{#1}   \fi
\ifx \showURL      \undefined \def \showURL       {\relax}        \fi
\providecommand\bibfield[2]{#2}
\providecommand\bibinfo[2]{#2}
\providecommand\natexlab[1]{#1}
\providecommand\showeprint[2][]{arXiv:#2}

\bibitem[Cai et~al\mbox{.}(2023)]{cai2024smplerx}
\bibfield{author}{\bibinfo{person}{Zhongang Cai}, \bibinfo{person}{Wanqi Yin},
  \bibinfo{person}{Ailing Zeng}, \bibinfo{person}{Chen Wei},
  \bibinfo{person}{Qingping Sun}, \bibinfo{person}{Yanjun Wang},
  \bibinfo{person}{Hui~En Pang}, \bibinfo{person}{Haiyi Mei},
  \bibinfo{person}{Mingyuan Zhang}, \bibinfo{person}{Lei Zhang},
  \bibinfo{person}{Chen~Change Loy}, \bibinfo{person}{Lei Yang}, {and}
  \bibinfo{person}{Ziwei Liu}.} \bibinfo{year}{2023}\natexlab{}.
\newblock \showarticletitle{{SMPLer-X}: Scaling Up Expressive Human Pose and
  Shape Estimation}. In \bibinfo{booktitle}{\emph{Advances in Neural
  Information Processing Systems}}, Vol.~\bibinfo{volume}{36}.
  \bibinfo{pages}{11454--11468}.
\newblock
\urldef\tempurl\url{https://doi.org/10.52202/075280-0506}
\showDOI{\tempurl}

\bibitem[Chen et~al\mbox{.}(2025)]{chen2025motion2motion}
\bibfield{author}{\bibinfo{person}{Ling-Hao Chen}, \bibinfo{person}{Yuhong
  Zhang}, \bibinfo{person}{Zixin Yin}, \bibinfo{person}{Zhiyang Dou},
  \bibinfo{person}{Xin Chen}, \bibinfo{person}{Jingbo Wang},
  \bibinfo{person}{Taku Komura}, {and} \bibinfo{person}{Lei Zhang}.}
  \bibinfo{year}{2025}\natexlab{}.
\newblock \showarticletitle{Motion2Motion: Cross-topology Motion Transfer with
  Sparse Correspondence}. In \bibinfo{booktitle}{\emph{Proceedings of the
  SIGGRAPH Asia 2025 Conference Papers}}. \bibinfo{publisher}{ACM},
  \bibinfo{pages}{1--11}.
\newblock
\urldef\tempurl\url{https://doi.org/10.1145/3757377.3763811}
\showDOI{\tempurl}

\bibitem[Choi et~al\mbox{.}(2021)]{choi2021tcmr}
\bibfield{author}{\bibinfo{person}{Hongsuk Choi}, \bibinfo{person}{Gyeongsik
  Moon}, \bibinfo{person}{Ju~Yong Chang}, {and} \bibinfo{person}{Kyoung~Mu
  Lee}.} \bibinfo{year}{2021}\natexlab{}.
\newblock \showarticletitle{Beyond Static Features for Temporally Consistent 3D
  Human Pose and Shape from a Video}. In \bibinfo{booktitle}{\emph{CVPR}}.
  \bibinfo{pages}{1964--1973}.
\newblock

\bibitem[Deng et~al\mbox{.}(2025)]{deng2025anymate}
\bibfield{author}{\bibinfo{person}{Yufan Deng}, \bibinfo{person}{Yuhao Zhang},
  \bibinfo{person}{Chen Geng}, \bibinfo{person}{Shangzhe Wu}, {and}
  \bibinfo{person}{Jiajun Wu}.} \bibinfo{year}{2025}\natexlab{}.
\newblock \showarticletitle{Anymate: A Dataset and Baselines for Learning 3D
  Object Rigging}. In \bibinfo{booktitle}{\emph{Proceedings of the SIGGRAPH
  2025 Conference Papers}}. \bibinfo{pages}{1--10}.
\newblock
\urldef\tempurl\url{https://doi.org/10.1145/3721238.3730743}
\showDOI{\tempurl}

\bibitem[Dwivedi et~al\mbox{.}(2024)]{dwivedi2024tokenhmr}
\bibfield{author}{\bibinfo{person}{Sai~Kumar Dwivedi}, \bibinfo{person}{Yu
  Sun}, \bibinfo{person}{Priyanka Patel}, \bibinfo{person}{Yao Feng}, {and}
  \bibinfo{person}{Michael~J. Black}.} \bibinfo{year}{2024}\natexlab{}.
\newblock \showarticletitle{{TokenHMR}: Advancing Human Mesh Recovery with a
  Tokenized Pose Representation}. In \bibinfo{booktitle}{\emph{Proceedings of
  the IEEE/CVF Conference on Computer Vision and Pattern Recognition (CVPR)}}.
  \bibinfo{pages}{1323--1333}.
\newblock

\bibitem[Gat et~al\mbox{.}(2025)]{gat2025anytop}
\bibfield{author}{\bibinfo{person}{Inbar Gat}, \bibinfo{person}{Sigal Raab},
  \bibinfo{person}{Guy Tevet}, \bibinfo{person}{Yuval Reshef},
  \bibinfo{person}{Amit~H. Bermano}, {and} \bibinfo{person}{Daniel Cohen-Or}.}
  \bibinfo{year}{2025}\natexlab{}.
\newblock \showarticletitle{{AnyTop}: Character Animation Diffusion with Any
  Topology}. In \bibinfo{booktitle}{\emph{Proceedings of the SIGGRAPH 2025
  Conference Papers}}. \bibinfo{publisher}{Association for Computing
  Machinery}, \bibinfo{pages}{1--10}.
\newblock
\urldef\tempurl\url{https://doi.org/10.1145/3721238.3730621}
\showDOI{\tempurl}

\bibitem[Goel et~al\mbox{.}(2023)]{goel2023humans}
\bibfield{author}{\bibinfo{person}{Shubham Goel}, \bibinfo{person}{Georgios
  Pavlakos}, \bibinfo{person}{Jathushan Rajasegaran}, \bibinfo{person}{Angjoo
  Kanazawa}, {and} \bibinfo{person}{Jitendra Malik}.}
  \bibinfo{year}{2023}\natexlab{}.
\newblock \showarticletitle{Humans in 4{D}: Reconstructing and Tracking Humans
  with Transformers}. In \bibinfo{booktitle}{\emph{Proceedings of the IEEE/CVF
  International Conference on Computer Vision (ICCV)}}.
  \bibinfo{pages}{14783--14794}.
\newblock

\bibitem[Gong et~al\mbox{.}(2026a)]{gong2026mocapanything}
\bibfield{author}{\bibinfo{person}{Kehong Gong}, \bibinfo{person}{Zhengyu Wen},
  \bibinfo{person}{Weixia He}, \bibinfo{person}{Mingxi Xu}, \bibinfo{person}{Qi
  Wang}, \bibinfo{person}{Ning Zhang}, \bibinfo{person}{Zhengyu Li},
  \bibinfo{person}{Dongze Lian}, \bibinfo{person}{Wei Zhao},
  \bibinfo{person}{Xiaoyu He}, {and} \bibinfo{person}{Mingyuan Zhang}.}
  \bibinfo{year}{2026}\natexlab{a}.
\newblock \showarticletitle{{MoCapAnything}: Unified {3D} Motion Capture for
  Arbitrary Skeletons from Monocular Videos}. In
  \bibinfo{booktitle}{\emph{Proceedings of the IEEE/CVF Conference on Computer
  Vision and Pattern Recognition (CVPR)}}. \bibinfo{pages}{7089--7099}.
\newblock
\urldef\tempurl\url{https://openaccess.thecvf.com/content/CVPR2026/html/Gong_MoCapAnything_Unified_3D_Motion_Capture_for_Arbitrary_Skeletons_from_Monocular_CVPR_2026_paper.html}
\showURL{\tempurl}

\bibitem[Gong et~al\mbox{.}(2026b)]{gong2026mocapanythingv2}
\bibfield{author}{\bibinfo{person}{Kehong Gong}, \bibinfo{person}{Zhengyu Wen},
  \bibinfo{person}{Dao~Thien Phong}, \bibinfo{person}{Mingxi Xu},
  \bibinfo{person}{Weixia He}, \bibinfo{person}{Qi Wang}, \bibinfo{person}{Ning
  Zhang}, \bibinfo{person}{Zhengyu Li}, \bibinfo{person}{Guanli Hou},
  \bibinfo{person}{Dongze Lian}, \bibinfo{person}{Xiaoyu He},
  \bibinfo{person}{Mingyuan Zhang}, {and} \bibinfo{person}{Hanwang Zhang}.}
  \bibinfo{year}{2026}\natexlab{b}.
\newblock \bibinfo{title}{{MoCapAnything V2}: End-to-End Motion Capture for
  Arbitrary Skeletons}.
\newblock
\showeprint[arxiv]{2604.28130}
\urldef\tempurl\url{https://arxiv.org/abs/2604.28130}
\showURL{\tempurl}
\newblock
\shownote{Accepted to ACM Transactions on Graphics (SIGGRAPH Asia 2026)}.

\bibitem[Kanazawa et~al\mbox{.}(2018)]{kanazawa2018hmr}
\bibfield{author}{\bibinfo{person}{Angjoo Kanazawa},
  \bibinfo{person}{Michael~J. Black}, \bibinfo{person}{David~W. Jacobs}, {and}
  \bibinfo{person}{Jitendra Malik}.} \bibinfo{year}{2018}\natexlab{}.
\newblock \showarticletitle{End-to-end Recovery of Human Shape and Pose}. In
  \bibinfo{booktitle}{\emph{CVPR}}. \bibinfo{pages}{7122--7131}.
\newblock

\bibitem[Kocabas et~al\mbox{.}(2020)]{kocabas2020vibe}
\bibfield{author}{\bibinfo{person}{Muhammed Kocabas}, \bibinfo{person}{Nikos
  Athanasiou}, {and} \bibinfo{person}{Michael~J Black}.}
  \bibinfo{year}{2020}\natexlab{}.
\newblock \showarticletitle{VIBE: Video inference for human body pose and shape
  estimation}. In \bibinfo{booktitle}{\emph{Proceedings of the IEEE/CVF
  conference on computer vision and pattern recognition}}.
  \bibinfo{pages}{5253--5263}.
\newblock

\bibitem[Kocabas et~al\mbox{.}(2021)]{kocabas2021pare}
\bibfield{author}{\bibinfo{person}{Muhammed Kocabas},
  \bibinfo{person}{Chun-Hao~P. Huang}, \bibinfo{person}{Otmar Hilliges}, {and}
  \bibinfo{person}{Michael~J. Black}.} \bibinfo{year}{2021}\natexlab{}.
\newblock \showarticletitle{PARE: Part Attention Regressor for 3D Human Body
  Estimation}. In \bibinfo{booktitle}{\emph{ICCV}}.
  \bibinfo{pages}{11127--11137}.
\newblock

\bibitem[Kolotouros et~al\mbox{.}(2019)]{kolotouros2019spin}
\bibfield{author}{\bibinfo{person}{Nikos Kolotouros}, \bibinfo{person}{Georgios
  Pavlakos}, \bibinfo{person}{Michael~J. Black}, {and} \bibinfo{person}{Kostas
  Daniilidis}.} \bibinfo{year}{2019}\natexlab{}.
\newblock \showarticletitle{Learning to Reconstruct 3D Human Pose and Shape via
  Model-fitting in the Loop}. In \bibinfo{booktitle}{\emph{ICCV}}.
  \bibinfo{pages}{2252--2261}.
\newblock

\bibitem[Li et~al\mbox{.}(2021b)]{li2021hybrik}
\bibfield{author}{\bibinfo{person}{Jiefeng Li}, \bibinfo{person}{Chao Xu},
  \bibinfo{person}{Zhicun Chen}, \bibinfo{person}{Siyuan Bian},
  \bibinfo{person}{Lixin Yang}, {and} \bibinfo{person}{Cewu Lu}.}
  \bibinfo{year}{2021}\natexlab{b}.
\newblock \showarticletitle{HybrIK: A Hybrid Analytical-Neural Inverse
  Kinematics Solution for 3D Human Pose and Shape Estimation}. In
  \bibinfo{booktitle}{\emph{CVPR}}. \bibinfo{pages}{3383--3393}.
\newblock

\bibitem[Li et~al\mbox{.}(2021a)]{li2021nbs}
\bibfield{author}{\bibinfo{person}{Peizhuo Li}, \bibinfo{person}{Kfir Aberman},
  \bibinfo{person}{Rana Hanocka}, \bibinfo{person}{Libin Liu},
  \bibinfo{person}{Olga Sorkine-Hornung}, {and} \bibinfo{person}{Baoquan
  Chen}.} \bibinfo{year}{2021}\natexlab{a}.
\newblock \showarticletitle{Learning Skeletal Articulations with Neural Blend
  Shapes}.
\newblock \bibinfo{journal}{\emph{ACM Transactions on Graphics}}
  \bibinfo{volume}{40}, \bibinfo{number}{4} (\bibinfo{year}{2021}),
  \bibinfo{pages}{1--15}.
\newblock
\urldef\tempurl\url{https://doi.org/10.1145/3450626.3459852}
\showDOI{\tempurl}

\bibitem[Li et~al\mbox{.}(2023)]{li2023genmm}
\bibfield{author}{\bibinfo{person}{Weiyu Li}, \bibinfo{person}{Xuelin Chen},
  \bibinfo{person}{Peizhuo Li}, \bibinfo{person}{Olga Sorkine-Hornung}, {and}
  \bibinfo{person}{Baoquan Chen}.} \bibinfo{year}{2023}\natexlab{}.
\newblock \showarticletitle{Example-Based Motion Synthesis via Generative
  Motion Matching}.
\newblock \bibinfo{journal}{\emph{ACM Transactions on Graphics}}
  \bibinfo{volume}{42}, \bibinfo{number}{4}, Article \bibinfo{articleno}{94}
  (\bibinfo{year}{2023}), \bibinfo{numpages}{12}~pages.
\newblock
\urldef\tempurl\url{https://doi.org/10.1145/3592395}
\showDOI{\tempurl}

\bibitem[Li et~al\mbox{.}(2022)]{li2022cliff}
\bibfield{author}{\bibinfo{person}{Zhihao Li}, \bibinfo{person}{Jianzhuang
  Liu}, \bibinfo{person}{Zhensong Zhang}, \bibinfo{person}{Songcen Xu}, {and}
  \bibinfo{person}{Youliang Yan}.} \bibinfo{year}{2022}\natexlab{}.
\newblock \showarticletitle{CLIFF: Carrying Location Information in Full Frames
  into Human Pose and Shape Estimation}. In \bibinfo{booktitle}{\emph{ECCV}}.
\newblock

\bibitem[Lin et~al\mbox{.}(2023)]{osx}
\bibfield{author}{\bibinfo{person}{Jing Lin}, \bibinfo{person}{Ailing Zeng},
  \bibinfo{person}{Haoqian Wang}, \bibinfo{person}{Lei Zhang}, {and}
  \bibinfo{person}{Yu Li}.} \bibinfo{year}{2023}\natexlab{}.
\newblock \showarticletitle{One-Stage {3D} Whole-Body Mesh Recovery with
  Component Aware Transformer}. In \bibinfo{booktitle}{\emph{Proceedings of the
  IEEE/CVF Conference on Computer Vision and Pattern Recognition (CVPR)}}.
  \bibinfo{pages}{21159--21168}.
\newblock
\urldef\tempurl\url{https://openaccess.thecvf.com/content/CVPR2023/html/Lin_One-Stage_3D_Whole-Body_Mesh_Recovery_With_Component_Aware_Transformer_CVPR_2023_paper.html}
\showURL{\tempurl}

\bibitem[Liu et~al\mbox{.}(2025)]{liu2025riganything}
\bibfield{author}{\bibinfo{person}{Isabella Liu}, \bibinfo{person}{Zhan Xu},
  \bibinfo{person}{Yifan Wang}, \bibinfo{person}{Hao Tan},
  \bibinfo{person}{Zexiang Xu}, \bibinfo{person}{Xiaolong Wang},
  \bibinfo{person}{Hao Su}, {and} \bibinfo{person}{Zifan Shi}.}
  \bibinfo{year}{2025}\natexlab{}.
\newblock \showarticletitle{RigAnything: Template-Free Autoregressive Rigging
  for Diverse 3D Assets}.
\newblock \bibinfo{journal}{\emph{ACM Transactions on Graphics}}
  \bibinfo{volume}{44}, \bibinfo{number}{4} (\bibinfo{year}{2025}),
  \bibinfo{pages}{1--12}.
\newblock
\urldef\tempurl\url{https://doi.org/10.1145/3731149}
\showDOI{\tempurl}

\bibitem[Loper et~al\mbox{.}(2015)]{loper2015smpl}
\bibfield{author}{\bibinfo{person}{Matthew Loper}, \bibinfo{person}{Naureen
  Mahmood}, \bibinfo{person}{Javier Romero}, \bibinfo{person}{Gerard
  Pons-Moll}, {and} \bibinfo{person}{Michael~J Black}.}
  \bibinfo{year}{2015}\natexlab{}.
\newblock \showarticletitle{{SMPL}: A Skinned Multi-Person Linear Model}.
\newblock \bibinfo{journal}{\emph{ACM Transactions on Graphics (TOG)}}
  \bibinfo{volume}{34}, \bibinfo{number}{6}, Article \bibinfo{articleno}{248}
  (\bibinfo{year}{2015}), \bibinfo{numpages}{16}~pages.
\newblock
\urldef\tempurl\url{https://doi.org/10.1145/2816795.2818013}
\showDOI{\tempurl}

\bibitem[Loshchilov and Hutter(2019)]{adamw}
\bibfield{author}{\bibinfo{person}{Ilya Loshchilov} {and}
  \bibinfo{person}{Frank Hutter}.} \bibinfo{year}{2019}\natexlab{}.
\newblock \showarticletitle{Decoupled Weight Decay Regularization}. In
  \bibinfo{booktitle}{\emph{International Conference on Learning
  Representations}}.
\newblock

\bibitem[Oquab et~al\mbox{.}(2024)]{oquab2024dinov2}
\bibfield{author}{\bibinfo{person}{Maxime Oquab}, \bibinfo{person}{Timoth{\'e}e
  Darcet}, \bibinfo{person}{Th{\'e}o Moutakanni}, \bibinfo{person}{Huy~V. Vo},
  \bibinfo{person}{Marc Szafraniec}, \bibinfo{person}{Vasil Khalidov},
  \bibinfo{person}{Pierre Fernandez}, \bibinfo{person}{Daniel Haziza},
  \bibinfo{person}{Francisco Massa}, \bibinfo{person}{Alaaeldin El-Nouby},
  \bibinfo{person}{Mahmoud Assran}, \bibinfo{person}{Nicolas Ballas},
  \bibinfo{person}{Wojciech Galuba}, \bibinfo{person}{Russell Howes},
  \bibinfo{person}{Po-Yao Huang}, \bibinfo{person}{Shang-Wen Li},
  \bibinfo{person}{Ishan Misra}, \bibinfo{person}{Michael Rabbat},
  \bibinfo{person}{Vasu Sharma}, \bibinfo{person}{Gabriel Synnaeve},
  \bibinfo{person}{Hu Xu}, \bibinfo{person}{Herv{\'e} J{\'e}gou},
  \bibinfo{person}{Julien Mairal}, \bibinfo{person}{Patrick Labatut},
  \bibinfo{person}{Armand Joulin}, {and} \bibinfo{person}{Piotr Bojanowski}.}
  \bibinfo{year}{2024}\natexlab{}.
\newblock \showarticletitle{{DINOv2}: Learning Robust Visual Features without
  Supervision}.
\newblock \bibinfo{journal}{\emph{Transactions on Machine Learning Research}}
  (\bibinfo{year}{2024}).
\newblock
\urldef\tempurl\url{https://openreview.net/forum?id=a68SUt6zFt}
\showURL{\tempurl}

\bibitem[Pang et~al\mbox{.}(2026)]{pang2026cast}
\bibfield{author}{\bibinfo{person}{Pu Pang}, \bibinfo{person}{Zhongfei Qing},
  \bibinfo{person}{Ling-Hao Chen}, \bibinfo{person}{Songnan Li},
  \bibinfo{person}{Hua Huang}, {and} \bibinfo{person}{Bo Hu}.}
  \bibinfo{year}{2026}\natexlab{}.
\newblock \bibinfo{title}{{CAST}: Real-Time Motion Capture for Any Skeleton
  Topology}.
\newblock
\showeprint[arxiv]{2610.14856}~[cs.CV]
\urldef\tempurl\url{https://arxiv.org/abs/2610.14856}
\showURL{\tempurl}

\bibitem[Pavlakos et~al\mbox{.}(2019)]{pavlakos2019expressive}
\bibfield{author}{\bibinfo{person}{Georgios Pavlakos},
  \bibinfo{person}{Vasileios Choutas}, \bibinfo{person}{Nima Ghorbani},
  \bibinfo{person}{Timo Bolkart}, \bibinfo{person}{Ahmed A.~A. Osman},
  \bibinfo{person}{Dimitrios Tzionas}, {and} \bibinfo{person}{Michael~J.
  Black}.} \bibinfo{year}{2019}\natexlab{}.
\newblock \showarticletitle{Expressive Body Capture: 3D Hands, Face, and Body
  from a Single Image}. In \bibinfo{booktitle}{\emph{CVPR}}.
  \bibinfo{pages}{10975--10985}.
\newblock

\bibitem[Radford et~al\mbox{.}(2021)]{clip}
\bibfield{author}{\bibinfo{person}{Alec Radford}, \bibinfo{person}{Jong~Wook
  Kim}, \bibinfo{person}{Chris Hallacy}, \bibinfo{person}{Aditya Ramesh},
  \bibinfo{person}{Gabriel Goh}, \bibinfo{person}{Sandhini Agarwal},
  \bibinfo{person}{Girish Sastry}, \bibinfo{person}{Amanda Askell},
  \bibinfo{person}{Pamela Mishkin}, \bibinfo{person}{Jack Clark},
  \bibinfo{person}{Gretchen Krueger}, {and} \bibinfo{person}{Ilya Sutskever}.}
  \bibinfo{year}{2021}\natexlab{}.
\newblock \showarticletitle{Learning Transferable Visual Models From Natural
  Language Supervision}. In \bibinfo{booktitle}{\emph{International Conference
  on Machine Learning}}. PMLR, \bibinfo{pages}{8748--8763}.
\newblock

\bibitem[Wu et~al\mbox{.}(2023a)]{wu2023dove}
\bibfield{author}{\bibinfo{person}{Shangzhe Wu}, \bibinfo{person}{Tomas Jakab},
  \bibinfo{person}{Christian Rupprecht}, {and} \bibinfo{person}{Andrea
  Vedaldi}.} \bibinfo{year}{2023}\natexlab{a}.
\newblock \showarticletitle{DOVE: Learning Deformable 3D Objects by Watching
  Videos}.
\newblock \bibinfo{journal}{\emph{International Journal of Computer Vision}}
  \bibinfo{volume}{131}, \bibinfo{number}{10} (\bibinfo{year}{2023}),
  \bibinfo{pages}{2623--2634}.
\newblock
\urldef\tempurl\url{https://doi.org/10.1007/s11263-023-01819-5}
\showDOI{\tempurl}

\bibitem[Wu et~al\mbox{.}(2023b)]{wu2023magicpony}
\bibfield{author}{\bibinfo{person}{Shangzhe Wu}, \bibinfo{person}{Ruining Li},
  \bibinfo{person}{Tomas Jakab}, \bibinfo{person}{Christian Rupprecht}, {and}
  \bibinfo{person}{Andrea Vedaldi}.} \bibinfo{year}{2023}\natexlab{b}.
\newblock \showarticletitle{{MagicPony}: Learning Articulated {3D} Animals in
  the Wild}. In \bibinfo{booktitle}{\emph{Proceedings of the IEEE/CVF
  Conference on Computer Vision and Pattern Recognition}}.
  \bibinfo{pages}{8792--8802}.
\newblock

\bibitem[Xu et~al\mbox{.}(2020)]{xu2020rignet}
\bibfield{author}{\bibinfo{person}{Zhan Xu}, \bibinfo{person}{Yang Zhou},
  \bibinfo{person}{Evangelos Kalogerakis}, \bibinfo{person}{Chris Landreth},
  {and} \bibinfo{person}{Karan Singh}.} \bibinfo{year}{2020}\natexlab{}.
\newblock \showarticletitle{{RigNet}: Neural Rigging for Articulated
  Characters}.
\newblock \bibinfo{journal}{\emph{ACM Transactions on Graphics (SIGGRAPH)}}
  \bibinfo{volume}{39}, \bibinfo{number}{4} (\bibinfo{year}{2020}).
\newblock

\bibitem[Xu et~al\mbox{.}(2022)]{xu2022morig}
\bibfield{author}{\bibinfo{person}{Zhan Xu}, \bibinfo{person}{Yang Zhou},
  \bibinfo{person}{Li Yi}, {and} \bibinfo{person}{Evangelos Kalogerakis}.}
  \bibinfo{year}{2022}\natexlab{}.
\newblock \showarticletitle{MoRig: Motion-Aware Rigging of Character Meshes
  from Point Clouds}. In \bibinfo{booktitle}{\emph{SIGGRAPH Asia 2022
  Conference Papers}}. \bibinfo{pages}{1--9}.
\newblock
\urldef\tempurl\url{https://doi.org/10.1145/3550469.3555390}
\showDOI{\tempurl}

\bibitem[Yang et~al\mbox{.}(2021)]{yang2021lasr}
\bibfield{author}{\bibinfo{person}{Gengshan Yang}, \bibinfo{person}{Deqing
  Sun}, \bibinfo{person}{Varun Jampani}, \bibinfo{person}{Daniel Vlasic},
  \bibinfo{person}{Forrester Cole}, \bibinfo{person}{Huiwen Chang},
  \bibinfo{person}{Deva Ramanan}, \bibinfo{person}{William~T. Freeman}, {and}
  \bibinfo{person}{Ce Liu}.} \bibinfo{year}{2021}\natexlab{}.
\newblock \showarticletitle{LASR: Learning Articulated Shape Reconstruction
  from a Monocular Video}. In \bibinfo{booktitle}{\emph{CVPR}}.
  \bibinfo{pages}{15980--15989}.
\newblock

\bibitem[Yang et~al\mbox{.}(2022)]{yang2022banmo}
\bibfield{author}{\bibinfo{person}{Gengshan Yang}, \bibinfo{person}{Minh Vo},
  \bibinfo{person}{Natalia Neverova}, \bibinfo{person}{Deva Ramanan},
  \bibinfo{person}{Andrea Vedaldi}, {and} \bibinfo{person}{Hanbyul Joo}.}
  \bibinfo{year}{2022}\natexlab{}.
\newblock \showarticletitle{BANMo: Building Animatable 3D Neural Models from
  Many Casual Videos}. In \bibinfo{booktitle}{\emph{CVPR}}.
  \bibinfo{pages}{2863--2873}.
\newblock

\bibitem[Yang et~al\mbox{.}(2023)]{yang2023synbody}
\bibfield{author}{\bibinfo{person}{Zhitao Yang}, \bibinfo{person}{Zhongang
  Cai}, \bibinfo{person}{Haiyi Mei}, \bibinfo{person}{Shuai Liu},
  \bibinfo{person}{Zhaoxi Chen}, \bibinfo{person}{Weiye Xiao},
  \bibinfo{person}{Yukun Wei}, \bibinfo{person}{Zhongfei Qing},
  \bibinfo{person}{Chen Wei}, \bibinfo{person}{Bo Dai}, \bibinfo{person}{Wayne
  Wu}, \bibinfo{person}{Chen Qian}, \bibinfo{person}{Dahua Lin},
  \bibinfo{person}{Ziwei Liu}, {and} \bibinfo{person}{Lei Yang}.}
  \bibinfo{year}{2023}\natexlab{}.
\newblock \showarticletitle{{SynBody}: Synthetic Dataset with Layered Human
  Models for {3D} Human Perception and Modeling}. In
  \bibinfo{booktitle}{\emph{Proceedings of the IEEE/CVF International
  Conference on Computer Vision (ICCV)}}. \bibinfo{pages}{20282--20292}.
\newblock
\urldef\tempurl\url{https://synbody.github.io/}
\showURL{\tempurl}

\bibitem[Yin et~al\mbox{.}(2026)]{yin2025smplestx}
\bibfield{author}{\bibinfo{person}{Wanqi Yin}, \bibinfo{person}{Zhongang Cai},
  \bibinfo{person}{Ruisi Wang}, \bibinfo{person}{Ailing Zeng},
  \bibinfo{person}{Chen Wei}, \bibinfo{person}{Qingping Sun},
  \bibinfo{person}{Haiyi Mei}, \bibinfo{person}{Yanjun Wang},
  \bibinfo{person}{Hui~En Pang}, \bibinfo{person}{Mingyuan Zhang},
  \bibinfo{person}{Lei Zhang}, \bibinfo{person}{Chen~Change Loy},
  \bibinfo{person}{Atsushi Yamashita}, \bibinfo{person}{Lei Yang}, {and}
  \bibinfo{person}{Ziwei Liu}.} \bibinfo{year}{2026}\natexlab{}.
\newblock \showarticletitle{SMPLest-X: Ultimate Scaling for Expressive Human
  Pose and Shape Estimation}.
\newblock \bibinfo{journal}{\emph{IEEE Transactions on Pattern Analysis and
  Machine Intelligence}} \bibinfo{volume}{48}, \bibinfo{number}{2}
  (\bibinfo{year}{2026}), \bibinfo{pages}{1778--1794}.
\newblock
\urldef\tempurl\url{https://doi.org/10.1109/TPAMI.2025.3618174}
\showDOI{\tempurl}

\bibitem[Zhang et~al\mbox{.}(2021)]{zhang2021pymaf}
\bibfield{author}{\bibinfo{person}{Hongwen Zhang}, \bibinfo{person}{Yating
  Tian}, \bibinfo{person}{Xinchi Zhou}, \bibinfo{person}{Wanli Ouyang},
  \bibinfo{person}{Yebin Liu}, \bibinfo{person}{Limin Wang}, {and}
  \bibinfo{person}{Zhenan Sun}.} \bibinfo{year}{2021}\natexlab{}.
\newblock \showarticletitle{PyMAF: 3D Human Pose and Shape Regression with
  Pyramidal Mesh Alignment Feedback Loop}. In \bibinfo{booktitle}{\emph{ICCV}}.
  \bibinfo{pages}{11446--11456}.
\newblock

\bibitem[Zhang et~al\mbox{.}(2025)]{zhang2025unirig}
\bibfield{author}{\bibinfo{person}{Jia-Peng Zhang}, \bibinfo{person}{Cheng-Feng
  Pu}, \bibinfo{person}{Meng-Hao Guo}, \bibinfo{person}{Yan-Pei Cao}, {and}
  \bibinfo{person}{Shi-Min Hu}.} \bibinfo{year}{2025}\natexlab{}.
\newblock \showarticletitle{One Model to Rig Them All: Diverse Skeleton Rigging
  with UniRig}.
\newblock \bibinfo{journal}{\emph{ACM Transactions on Graphics}}
  \bibinfo{volume}{44}, \bibinfo{number}{4}, Article \bibinfo{articleno}{123}
  (\bibinfo{year}{2025}), \bibinfo{numpages}{18}~pages.
\newblock
\urldef\tempurl\url{https://doi.org/10.1145/3730930}
\showDOI{\tempurl}

\bibitem[Zhang et~al\mbox{.}(2026)]{zhang2026skintokens}
\bibfield{author}{\bibinfo{person}{Jia-peng Zhang}, \bibinfo{person}{Cheng-Feng
  Pu}, \bibinfo{person}{Meng-Hao Guo}, \bibinfo{person}{Yan-Pei Cao}, {and}
  \bibinfo{person}{Shi-Min Hu}.} \bibinfo{year}{2026}\natexlab{}.
\newblock \bibinfo{title}{Skin Tokens: A Learned Compact Representation for
  Unified Autoregressive Rigging}.
\newblock
\showeprint[arxiv]{2602.04805}
\urldef\tempurl\url{https://arxiv.org/abs/2602.04805}
\showURL{\tempurl}

\end{thebibliography}
\end{document}